\documentclass[letterpaper, 10 pt, conference]{ieeeconf}  

\IEEEoverridecommandlockouts                              

\usepackage{graphics} 
\usepackage{epsfig} 
\usepackage{graphicx}
\usepackage{epstopdf} 
\usepackage{mathptmx} 
\usepackage{times} 
\usepackage{amsmath}
\usepackage{amssymb}
\usepackage{microtype}
\usepackage{bbm}
\usepackage{dsfont}
\usepackage{subcaption}
\usepackage{algorithmic}
\usepackage{algorithm}
\usepackage{subcaption}
\usepackage{booktabs} 
\usepackage{multirow}
\usepackage{hyperref}
\hypersetup{breaklinks=true,colorlinks=true,linkcolor=blue,citecolor=blue}
\usepackage{color}

\newcommand{\vect}[1]{\mathbf{#1}}

\DeclareMathOperator{\argmin}{arg\,min}

\title{\LARGE \bf
Learning When to Use Adaptive Adversarial Image Perturbations against Autonomous Vehicles
}

\author{Hyung-Jin Yoon$^{{ 1}}$, Hamidreza Jafarnejadsani$^{2}$, and Petros Voulgaris$^{1}$
\thanks{*Research supported by NSF CPS \#1932529, NSF CMMI \#1663460,  NSF CMMI \#2137753 and UNR internal funding.}
\thanks{$^{1}$Hyung-Jin Yoon and Petros Voulgaris are with the Department of Mechanical Engineering, University of Nevada, Reno, NV 89557, USA
        {\tt\small \{hyungjiny, pvoulgaris\}@unr.edu}}%
\thanks{$^{2}$Hamidreza Jafarnejadsani is with the Department of Mechanical Engineering, Stevens Institute of Technology, NJ 07030, USA
        {\tt\small hjafarne@stevens.edu}}%
}

\begin{document}

\maketitle
\thispagestyle{plain}
\pagestyle{plain}

\begin{abstract}

Deep neural network (DNN) models are widely used in autonomous vehicles for object detection using camera images. However, these models are vulnerable to adversarial image perturbations. Existing methods for generating these perturbations use each incoming image frame as the decision variable, resulting in a computationally expensive optimization process that starts over for each new image. Few approaches have been developed for attacking online image streams while considering the physical dynamics of autonomous vehicles, their mission, and the environment. To address these challenges, we propose a multi-level stochastic optimization framework that monitors the attacker's capability to generate adversarial perturbations. Our framework introduces a binary decision attack/not attack based on the attacker's capability level to enhance its effectiveness. We evaluate our proposed framework using simulations for vision-guided autonomous vehicles and actual tests with a small indoor drone in an office environment. Our results demonstrate that our method is capable of generating real-time image attacks while monitoring the attacker's proficiency given state estimates.

\end{abstract}


\section{Introduction}
Machine learning (ML) tools that detect objects using high-dimensional sensors, such as camera images~\cite{redmon2016you} or point clouds measured by LiDAR~\cite{qi2017pointnet}, are extensively used in autonomous vehicles~\cite{DJI_MAVIC, waymo_data_challenge}. As vision-based autonomous vehicles become more integrated into society, it is crucial to ensure the robustness of these systems, which rely on various sensor signals in uncertain environments. Analyzing worst-case scenarios within uncertainties has been a useful approach to robustify control systems~\cite{bacsar2008h} and reinforcement learning~\cite{pinto2017robust}. To follow this approach, researchers have revealed the vulnerability of machine learning methods, especially deep learning tools developed for computer vision tasks such as object detection and classification, to data perturbed by adversaries. For instance, small perturbations can be added to images that are unnoticeable to human eyes but result in incorrect image classifications~\cite{kurakin2016adversarial_at_scale,kurakin2016adversarial_world,goodfellow2014explaining}.
Moreover, recent works have demonstrated adversarial image perturbations against autonomous vehicles, including (1) modifying physical objects, such as putting stickers on a road~\cite{ackerman2019three} or a road sign~\cite{eykholt2018robust}, to fool an ML image classifier or end-to-end vision-based autonomous car; and (2) fooling object tracking algorithms in autonomous driving systems~\cite{jia2020fooling}.
Adversarial machine learning commonly focuses on creating stealthy and natural-looking perturbations to evade human detection. Such attacks are designed to resemble out-of-distribution samples that may occur in real-world environments. As a consequence, ensuring the robustness of ML-based autonomous vehicle systems against adversarial attacks has become increasingly critical.

While the aforementioned adversarial image perturbations against autonomous cars~\cite{ackerman2019three, eykholt2018robust, jia2020fooling} successfully reveal weaknesses in vision-guided navigation in autonomous vehicles, these perturbed images are generated offline. However, offline methods~\cite{eykholt2018robust, jia2020fooling} do not consider the effect of real-time attacks on dynamically changing environments during driving or flight of the vehicles. To prevent accidents~\cite{tesla21crash} caused by vision-guided autonomous vehicles due to defective perception systems and their vulnerabilities, we need to study attack and defense techniques that go beyond offline methods for deep neural networks.

There are two approaches to generating adversarial image perturbations, depending on the attacker's access to the victim perception model. In the \emph{white-box} attack approach, the attacker has full access to the victim ML classifier (or object detector) and generates adversarial image perturbations through iterative optimization~\cite{kurakin2016adversarial_world, jia2020fooling}. In this method, images are the decision variables of optimization, and the training loss function is reused with incorrect labels set by the attacker. The optimization takes iterative gradient steps with respect to the image variables calculated using back-propagation through the known victim ML classifier~\cite{kurakin2016adversarial_world} (or object detector~\cite{jia2020fooling}). On the other hand, in the \emph{black-box} approaches~\cite{ilyas2018black, wei2020heuristic}, the attacker only has access to input and output pairs of the victim model and must estimate the gradient. However, estimating the gradients in \emph{black-box} attacks requires a large number of samples, which may not be available from autonomous systems operating in dynamic environments.

\begin{figure*}[!ht]
\centering
\subfloat[][Adversarial patch example. ]{\includegraphics[width=0.33\linewidth]{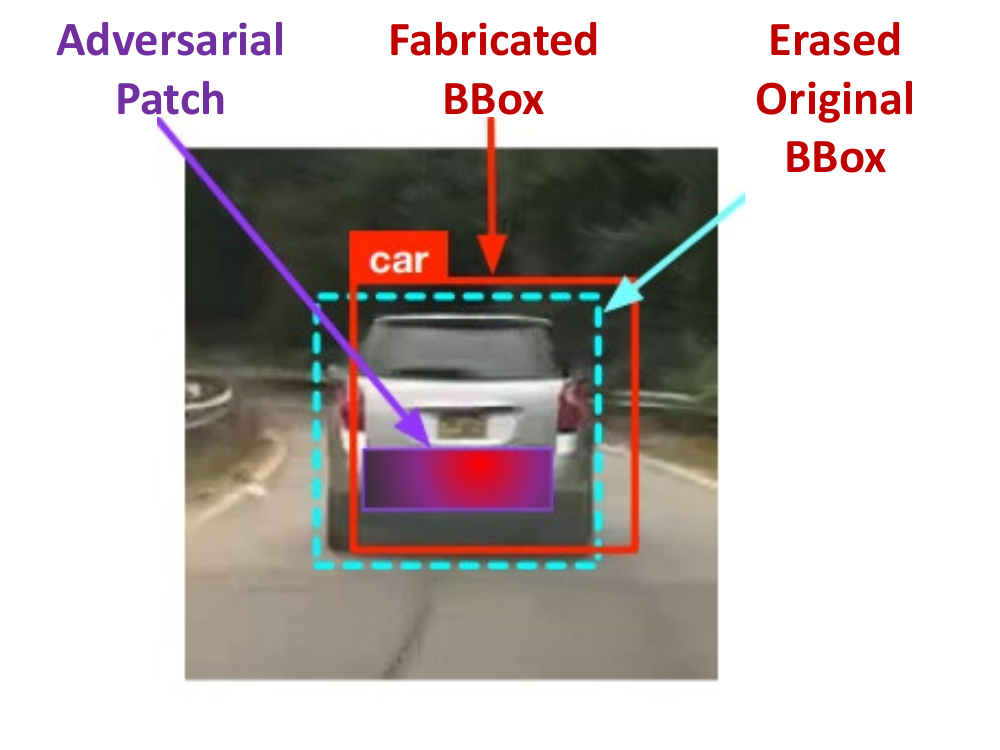}
}
\subfloat[][Choosing when to use the adversarial image perturbation .]{\includegraphics[width=0.66\linewidth]{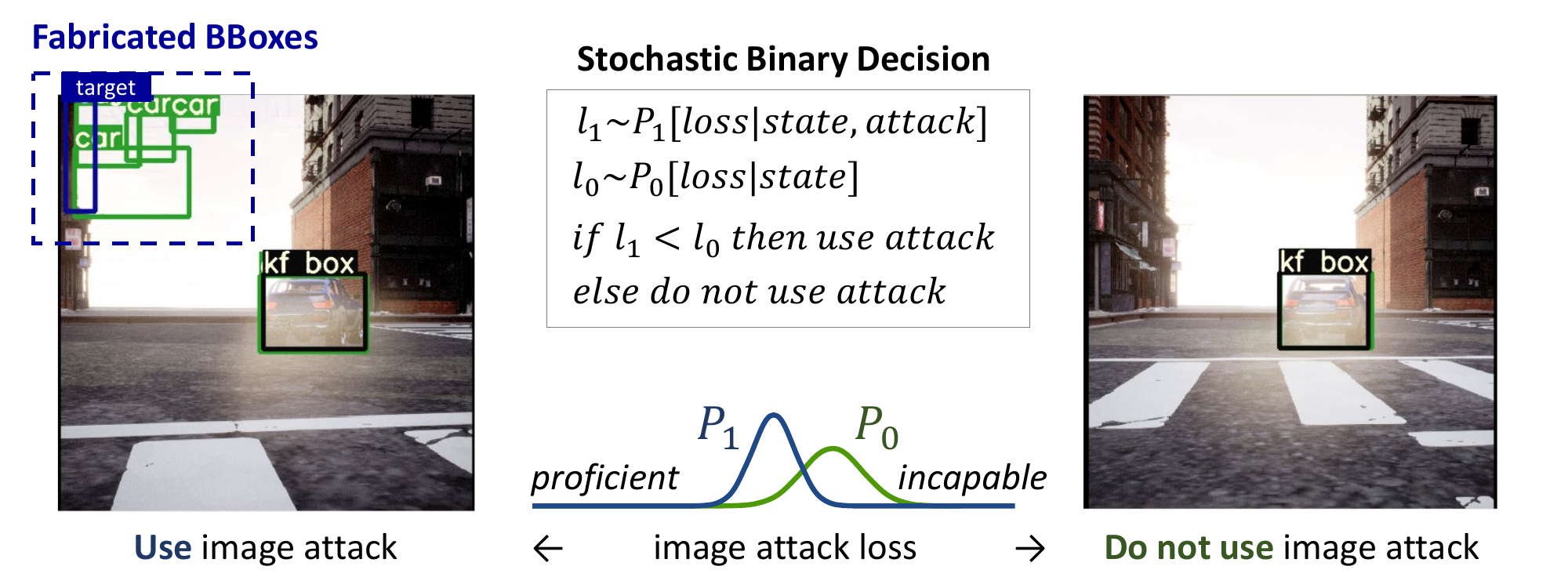}
}
\caption{Image attacks: (a) Adversarial patch in~\cite{jia2020fooling}, (b) Adversarial perturbation with binary decision in this paper. \emph{kf box} denotes Kalman filtered bounding box (BBox).}
\label{fig0_adv_img_attack}
\vskip -0.25 in 
\end{figure*}

{\it Statement of Contribution:} To our knowledge, this paper is the first to propose a stealthy attack scheme on image streams used for object detection/tracking in autonomous vehicles (e.g., self-driving cars and drones) that can be deployed {\it online}, and the {\it physical dynamics of the system} and the {\it varying surrounding environment} are taken into account in the optimization phase of the attack scheme. In this paper, we present a framework that utilizes generative adversarial networks (GANs) to generate adversarial images in real-time scenarios without the need for iterative steps. Building on the approach outlined in~\cite{xiao2018generating}, our proposed multi-level framework consists of several components. First, the GAN functions as an online image generator. Second, a reinforcement learning agent is trained to misguide the vehicle according to the adversary's objective. Lastly, a binary decision-maker determines when to use image attacks based on the proficiency of the image attack generator, given the current state estimate. Our framework provides a more efficient and practical alternative to iterative \emph{white-box} methods for generating adversarial images.

Our contributions can be summarized as follows:
\begin{itemize}
\item We propose a {\it real-time} adversarial image perturbation framework that allows for implementation on real-world robots, in contrast to existing offline methods.
\item We introduce a {\it state estimation-based reinforcement learning} approach that learns to decide on the image frame area to fabricate bounding boxes. This approach eliminates the need for manual annotation of patch areas.
\item We incorporate a constraint on the strength of the image perturbation, making the attacked image frame {\it less noticeable and more stealthy} compared to existing methods. This is demonstrated in Figure~\ref{fig0_adv_img_attack}.
\end{itemize}

\section{Related Works}
Adversarial image perturbations have been extensively studied to attack autonomous vehicles that rely on camera images for navigation~\cite{boloor2020attacking, jia2020fooling, jha2020ml}. For instance, in~\cite{boloor2020attacking}, an optimization problem was formulated to place black marks on the road, which caused an end-to-end autonomous driving car to veer off the road in a virtual reality environment. This approach was inspired by the demonstration of attacking Tesla's autonomous driving systems with just three small stickers~\cite{ackerman2019three}. In another work~\cite{jia2020fooling}, the authors demonstrated the effectiveness of a \emph{white-box} adversarial image perturbation method on object tracking of an autonomous system that uses \emph{Kalman} filter (KF) to disrupt the object tracking. This method was also shown to be effective in attacking an industry-level perception module that uses vision-based object detection fused with LIDAR, GPS, and IMU~\cite{jha2020ml}.

The aforementioned \emph{white-box} methods~\cite{jia2020fooling,jha2020ml} require full sets of iterative optimization computations for every new image, rendering them unsuitable for dynamic environments with evolving situations and control loops of autonomous vehicles. These approaches do not consider the varying computation time of the iterative optimizations, which might have different termination steps for online applications. Additionally, the attack methods in~\cite{jia2020fooling,jha2020ml} often require additional state information that is not always readily available, unlike the image stream. For instance, generating the adversarial patch in Figure~\ref{fig0_adv_img_attack}a in~\cite{jia2020fooling} requires the attacker to know the exact anchor index associated with the target bounding box (BBox) and the location to place the BBox. As mentioned in the \emph{open review}~\cite{open_review} by the authors in~\cite{jia2020fooling}, the adversarial patch area was manually annotated in each video frame.

Although there are various other adversarial image attack methods available, many of them are offline methods that require additional information such as labeled training data to generate adversarial images.

\section{Real-time Adversarial Image Attack}

Our goal is to develop a real-time solution that can learn to generate adversarial image perturbations and decide when to use the attack based on the proficiency of the attack generator, as illustrated in Figure~\ref{fig0_adv_img_attack}b. The adversarial image perturbations are designed to manipulate the perception of autonomous vehicles to misguide them according to the adversary's objectives, such as causing collisions or making the vehicle deviate from its original path. To formally formulate the problem, We consider the following assumptions and settings.

\subsection{Problem description and proposed framework}

\begin{figure}[ht]
\begin{center}
\centerline{\includegraphics[width=\columnwidth]{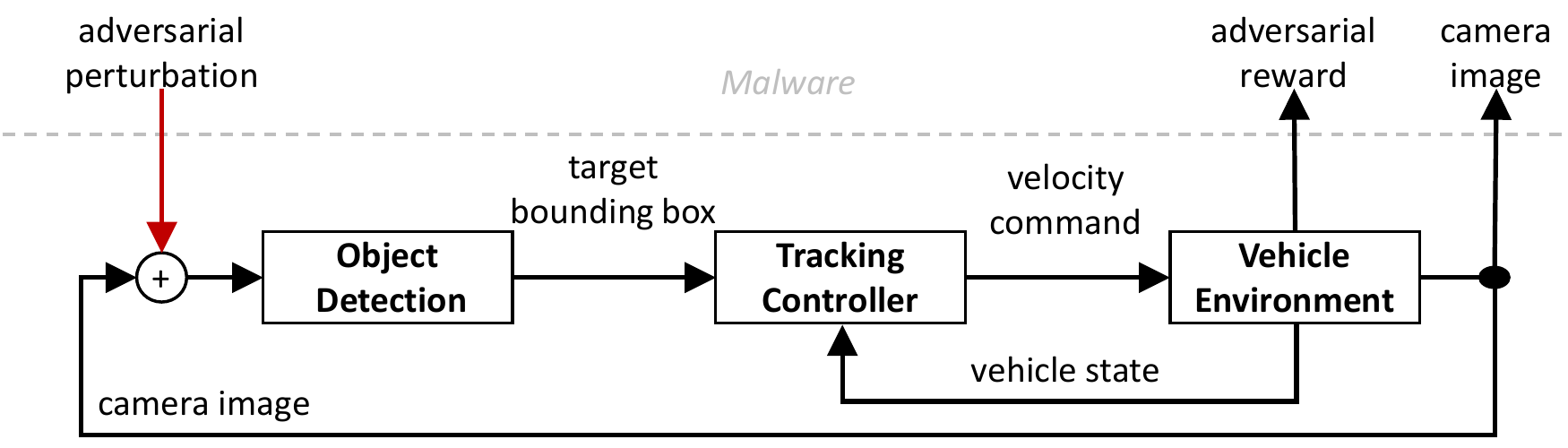}}
\caption{Attacker (malware) and victim system (guidance)}
\label{fig1_victim_sys_malware}
\end{center}
\vskip -0.25in
\end{figure}

We focus on an autonomous vehicle that utilizes an object detection ML method to track a target object using camera images, as shown in Figure~\ref{fig1_victim_sys_malware}. We used a recent version of the \emph{YOLO} object detection model~\cite{redmon2016you}, which was downloaded from~\cite{yolov5}, for our experiments\footnote{Another popular object detection model, \emph{Faster R-CNN}, can be attacked using similar White box attack method as in~\cite{wang2020adversarial}. Hence, our proposed method can be implemented with \emph{Faster R-CNN}.}. The output of the object detector network is a multi-dimensional tensor that is processed using non-max suppression~\cite{redmon2016you} to obtain a list of bounding box coordinates. The box with the highest confidence score for the target class is then selected from the list of detected bounding boxes to generate tracking control commands. The autonomous guidance system uses the vehicle's actuators, including the acceleration pedal, brake, and steering wheel, to keep the target's bounding box centered in the camera view and within a specified size range. Consequently, the vehicle moves towards and tracks the target object.

We assume the adversary's objective is to disrupt the target tracking control in Figure~\ref{fig1_victim_sys_malware}. The attacker is assumed to be embedded as \emph{Malware} and has access to the image stream, enabling them to perturb the input to the object detection module of the victim system, as illustrated in Figure~\ref{fig1_victim_sys_malware}. Given the image streams denoted as ${\mathbf{x}_0, \mathbf{x}_1, ..., \mathbf{x}_t}$, the attacker's goal is to generate adversarial image perturbations ${\vect{w}_0, \vect{w}_1, ..., \vect{w}_t}$ that mislead the victim vehicle according to adversarial objectives expressed in terms of adversarial rewards ${r_1, r_2, ..., r_t}$. The reward function is based on the vehicle's state, such as position, velocity, or collision states, and actions that involve the coordinates used to fabricate the bounding boxes through the image attack generator, as shown in Figure~\ref{fig2_proposed_framework}. These rewards are crucial for applying \emph{reinforcement learning} (RL), which learns the correlation between actions and rewards for different states of the system. In this framework, a binary decision-maker determines when to attack based on the attack proficiency (represented as loss in Figure~\ref{fig2_proposed_framework}). The problem addressed in this framework can be summarized as follows:

\begin{figure}[ht]
\vskip -0.05in
\begin{center}
\centerline{\includegraphics[width=\linewidth]{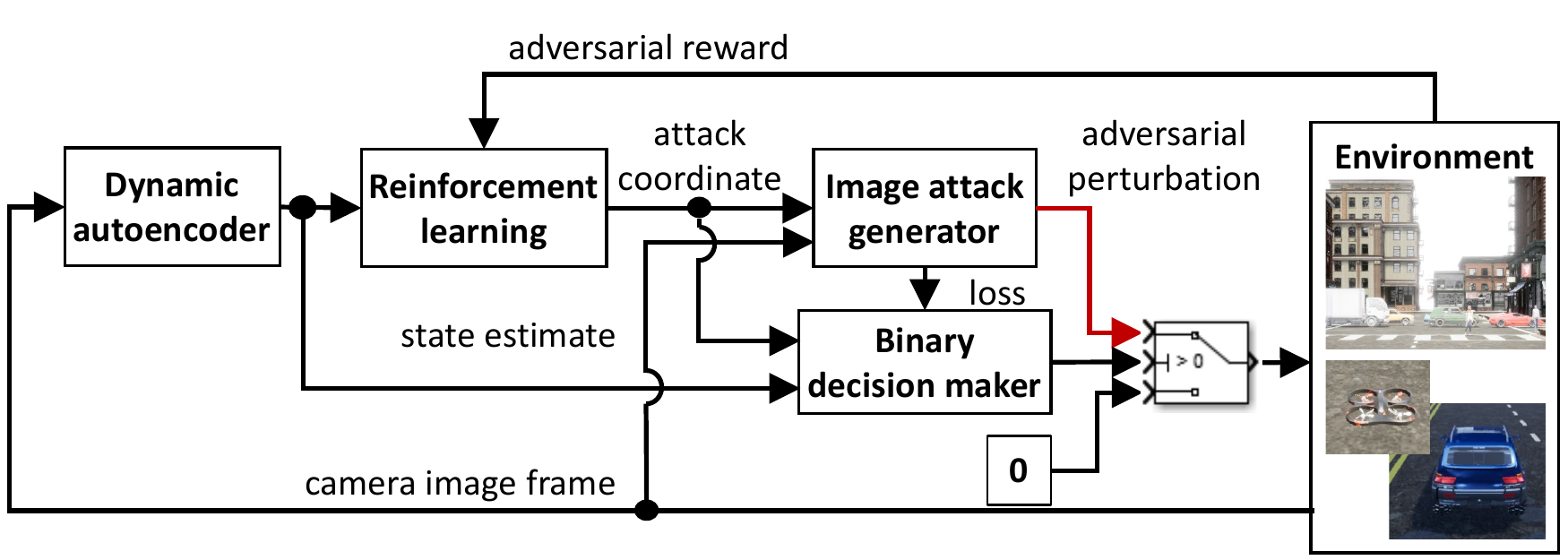}}
\caption{Image attack framework with binary decision maker.}
\label{fig2_proposed_framework}
\end{center}
\vskip -0.25in
\end{figure}

\noindent \textbf{Problem}:
Develop machine learning methods that learn to increase the sum of rewards $\{ r_t \}$ for the adversary by generating adversarial perturbation $\{ \vect{w}_t \}$ while selecting when to use the attack at the time step $t$, as shown in Figure~\ref{fig2_proposed_framework}. The ML method assumes to use only the image stream $\{ \vect{x}_t \}$ from the autonomous vehicle that has a guidance system and malware shown in Figure~\ref{fig1_victim_sys_malware}.  

\subsection{Online image attack with binary decision making}\label{sec:online_image_attack_w_switch}
Our framework involves binary decision-making that depends on the proficiency of the image attack generator. This type of decision-making belongs to the multi-armed bandit class of problems~\cite{vermorel2005multi}, where the decision-maker selects the most profitable action. However, unlike the classical multi-armed bandit, where the rewards are generated from independent-stationary distributions, our decision-maker must consider non-stationary system state distributions. Specifically, given the attack coordinate chosen by RL and the state estimate from the dynamic autoencoder, the decision-maker must determine whether using the attack is profitable or not. To tackle this challenge, the authors in~\cite{zhang2020neural} used a deep neural network (DNN) to learn the correlation between the state, decision, and profit. They also employed random dropout~\cite{gal2016dropout} with the DNN to estimate the profit distributions for each decision. This multi-armed bandit algorithm, which uses DNN with random dropout, is known as ~\emph{Neural Thompson Sampling} (NTS).

We sought to implement NTS for binary decision-making, using the proficiency of the image attack generator as the profits in the multi-armed bandit. While the direct application of NTS to our framework is appealing, there is a causality issue to consider. Specifically, the loss value is independent of binary decision-making, as it depends on the attack coordinates and the image frame. In our experiments, we tested NTS, but it did not demonstrate the desired behavior of selecting to attack when the expected loss value is low.

Therefore, we propose an alternative method to NTS that involves comparing two conditional expectations. Specifically, our method compares $E[l_t|\vect{h}_t, \vect{a}_t]$ with $E[l_t|\vect{h}_t]$. Here, $l_t$ represents the loss function used to measure the proficiency of the image attack GAN. The state estimates that are low-dimensional representations of all previous observations, denoted by $\vect{h}_t$,
are obtained using the dynamic autoencoder shown in Figure~\ref{fig2_proposed_framework}. Since the true states $\mathbf{s}_t$ are only partially observed through the image $\mathbf{x}_t$, $\vect{h}_t$ provides a better estimate of the state. Additionally, $\vect{a}_t$ represents the action determined by reinforcement learning agent in Figure~\ref{fig2_proposed_framework}. This action is the attack coordinate, which is the position and size of the fabricated bounding box. The goal of this approach is to compare the expected loss given the attack coordinate $\vect{a}_t$ suggested by RL with the expected loss averaged over all other possible attack coordinates. If the expected loss given $\vect{a}_t$ is lower than the average loss, then $\vect{a}_t$ is considered a promising attack coordinate to be used at this point. We refer to this decision-making method as \emph{Conditional Sampling} (CS). The loss estimation and decision-making procedure of CS are as follows:

\noindent\textbf{Estimation:} The estimation for CS involves the following optimizations:
{\small
\begin{equation}\label{eqn:attack_loss_prediction}
\begin{aligned}
    \argmin_{\theta^\text{dec}} &\Vert l_t - \hat{l}_0(\vect{h}_t;\theta^\text{dec}) \Vert^2, \\
    \argmin_{\theta^\text{dec}} &\Vert l_t - \hat{l}_1(\vect{h}_t, \vect{a}_t;\theta^\text{dec}) \Vert^2,
\end{aligned}
\end{equation}
}

\noindent where $\hat{l}_0$ and $\hat{l}_1$ are DNNs trained to predict the loss functions values $l_t$ given the state estimate $\vect{h}_t$ and the attack coordinate $\vect{a}_t$ respectively. The DNNs have parameters denoted as $\theta^\text{dec}$ that need to be optimized.

\noindent\textbf{Decision making:} 
The decision to launch an attack is determined by random sampling. To obtain sample image attack losses $\tilde{l}_0$ and $\tilde{l}_1$, we follow the same approach as in NTS, using the current state estimate $\vect{h}_t$ and the attack coordinate $\vect{a}_t$. Specifically, we generate output samples by applying random dropout in DNNs, i.e., $\hat{l}_0$ and $\hat{l}_1$, and estimate Gaussian distributions based on these samples. We then sample from the estimated Gaussian distributions to obtain $\tilde{l}_0$ and $\tilde{l}_1$, which can be expressed as:
{\small
\begin{equation}\label{eqn:CS}
\tilde{l}_0 \sim \hat{l}_0(\vect{h}_t;\theta^\text{dec}) \quad \text{and} \quad \tilde{l}_1 \sim \hat{l}_1(\vect{h}_t, \vect{a}_t;\theta^\text{dec}).
\end{equation}
}

\noindent Further details of the sampling procedure can be found in \cite{zhang2020neural}. To make a decision, we select the option with the lower loss value, i.e., if $\tilde{l}_0 < \tilde{l}_1$, then the attack will not be launched in the time step; otherwise, the image attack will be performed. The conditional probability distributions for the samplings are denoted as $P_0[l_t|\vect{h}_t]$ and $P_1[l_t|\vect{h}_t, \vect{a}_t]$, as shown in Figure~\ref{fig0_adv_img_attack}b.

\begin{figure}[ht]
\begin{center}
\centerline{\includegraphics[width=\columnwidth]{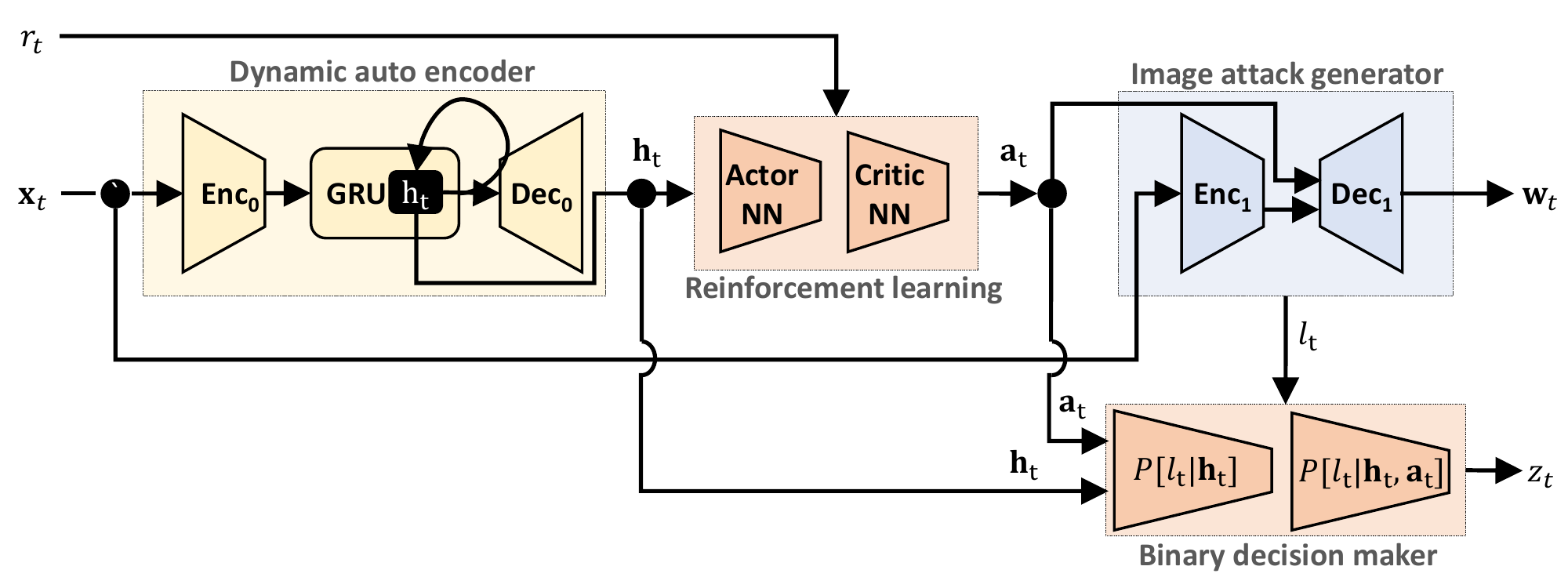}}
\caption{Multi-level image attack computation network. The computation network for multi-level image attack consists of image encoders and decoders within the dynamic autoencoder and image attack generator, which are adapted from~\cite{goodfellow2020generative}.}
\label{fig3_computation_network}
\end{center}
\vskip -0.25in
\end{figure}

The proposed framework integrates estimation models for binary decision-making within computation networks consisting of DNNs, as depicted in Figure~\ref{fig3_computation_network}. A major advantage of this framework is the ability to generate real-time adversarial image perturbations through recursive computations. The process involves feeding the image $\mathbf{x}_t$ at time $t$ into encoder networks, $\vect{Enc}_0$ and $\vect{Enc}_1$, for dimension reduction. $\vect{Enc}_0$ is used for state estimation, while $\vect{Enc}_1$ generates the perturbed image $\vect{w}_t$. The dynamic autoencoder comprises $\vect{Enc}_0$, $\vect{GRU}$ (gated recurrent unit), and $\vect{Dec}_0$ (decoder). The $\vect{GRU}$ recursively updates the hidden state $\vect{h}_t$ using the encoded image $\vect{Enc}_0(\vect{x}_t)$ and the high-level attack action $\vect{a}_t$, as shown in Figure~\ref{fig3_computation_network}. The estimated state information in $\mathbf{h}_t$ is then used by the actor (policy) to generate the high-level action, i.e., $\mathbf{a}_t = \vect{Actor}(\mathbf{h}_t)$.

As shown in Figure~\ref{fig3_computation_network}, the adversarial image perturbation $\vect{w}_t$ is generated by $\vect{Dec}_1$ using the high-level attack action $\mathbf{a}_t$ and another encoded image from $\vect{Enc}_1$, i.e., $\vect{w}_t = \vect{Dec}_1(\vect{Enc}_1(\mathbf{x}_t), \mathbf{a}_t)$. The perturbed image frame is obtained by applying the perturbation to the original image with a scale factor $\alpha$, i.e., $\tilde{\mathbf{x}}_t= \max(\min(\mathbf{x}_t + \alpha \mathbf{w}_t, 1),0)$.
The binary decision maker selects the decision variable $z_t$ using the conditional sampling (CS) described in~\eqref{eqn:CS}, where $z_t=1$ indicates that the attack is used and $z_t=0$ indicates that the attack is not used.

The recursive process of generating adversarial image perturbation using only camera image is summarized in Algorithm~\ref{alg:recursive_attack}. The entire computation at each time-step uses only the current observation or the state values in the previous time-step without iterative optimization, enabling real-time generation of image attacks.
{\small
\begin{algorithm}[ht]
   \caption{Recursive Image Attack}
   \label{alg:recursive_attack}
\begin{algorithmic}
   \STATE {\bfseries Initialize:} $ t \leftarrow 0$ ; load the pre-trained parameters of the recursive attack networks.
   \REPEAT
   \STATE Generate attack command using RL policy (Actor) 
   \STATE $\quad \mathbf{a}_t \leftarrow \vect{Actor}(\mathbf{h}_t)$
   \STATE Encode the observed image $\mathbf{x}_t$ from the environment
   \STATE $\quad \mathbf{\zeta}_t \leftarrow \vect{Enc}_1(\mathbf{x}_t)$
   \STATE Generate adversarial image perturbation
   \STATE $\quad \mathbf{w}_t \leftarrow \vect{Dec}_1(\vect{\zeta}_t, \mathbf{a}_t)$
   \STATE Feed $\mathbf{w}_t$ to the environment and get new image $\mathbf{x}_{t+1}$
   \STATE $\quad \mathbf{x}_{t+1}, \mathbf{s}_{t+1}, r_{t}, \text{done} \leftarrow \vect{Environment}(\vect{s}_t, \vect{w}_t)$
   \STATE Recursively update the state predictor $\vect{h}_{t+1}$ with $\vect{x}_{t+1}$
   \STATE $\quad \mathbf{h}_{t+1} \leftarrow \vect{GRU}(\mathbf{h}_t, \vect{Enc}_0(\vect{x}_{t+1}), \vect{a}_t)$
   \STATE Sample from the conditional distribution as in~\eqref{eqn:CS}, i.e.,
   \STATE $\quad l_0 \sim   P_0[l_t|\vect{h}_t] \quad \text{and} \quad l_1 \sim   P_1[l_t|\vect{h}_t, \vect{a}_t].$
   \STATE Use $\vect{w}_t$ if $l_0 < l_1$. Otherwise do not use it.
   \UNTIL{$\text{done}$ is True, i.e., the episode terminates with a terminal condition.}
\end{algorithmic}
\end{algorithm} 
}

\subsection{Multi-time scale optimization to train the attacker}\label{sec:multileveloptim}
We employ a multi-level stochastic optimization approach that separates the time scales of the updates for the various components depicted in Figure~\ref{fig2_proposed_framework}. Our stochastic optimization method trains the multi-level image attack computational networks illustrated in Figure~\ref{fig3_computation_network}. During training, the learning components and the environment are coupled and update their parameters simultaneously. The choice of time scales for the updates can have a significant impact on the behavior of the multi-time scale optimization process. For instance, in actor-critic~\cite{konda2000actor}, the critic has a faster update rate than the actor. In contrast, in the generative adversarial network described in~\cite{heusel2017gans}, the discriminator has a faster update rate than the generator. Following the heuristics and theories described in~\cite{konda2000actor, heusel2017gans}, we set slower parameter update rates for the lower-level components.

Let us denote the parameters of the various components as follows: $\theta_n^\text{img}$ represents the parameters of $\vect{Enc}_1(\cdot)$ and $\vect{Dec}_1(\cdot)$, $\theta_n^\text{sys}$ represents the parameters of the dynamic autoencoder comprising $\vect{Enc}_0(\cdot)$, $\vect{GRU}(\cdot)$, and $\vect{Dec}_0(\cdot)$, $\theta_n^\text{actor}$ represents the parameters of the actor denoted by $\vect{Actor}(\cdot)$, and $\theta_n^\text{critic}$ represents the parameters of the critic denoted by $Q(\cdot , \cdot)$, which is the action-value function for policy evaluation. We update these parameters using different step sizes, based on the pace of the update rates, as follows:
{\small
\begin{equation}\label{eq:multi-time-scale}
  \begin{aligned}
   \begin{matrix}
    \theta^\text{img}_{n+1}&=&\theta^\text{img}_{n}&+&\epsilon_n^\text{img}& S_n^\text{img}(\mathcal{M}_\text{trajectory})\\
    \theta^\text{dec}_{n+1}&=&\theta^\text{dec}_{n}&+&\epsilon_n^\text{dec}& S_n^\text{dec}(\mathcal{M}_\text{decision})\\
    \theta^\text{actor}_{n+1}&=&\theta^\text{actor}_{n}&+&\epsilon_n^\text{actor}& S_n^\text{actor}(\mathcal{M}_\text{transition})\\
    \theta^\text{critic}_{n+1}&=&\theta^\text{critic}_{n}&+&\epsilon_n^\text{critic}&
    S_n^\text{critic}(\mathcal{M}_\text{transition})\\
    \theta^\text{sys}_{n+1}&=&\theta^\text{sys}_{n}&+&\epsilon_n^\text{sys}& S_n^\text{sys}(\mathcal{M}_\text{trajectory})
    \end{matrix}
  \end{aligned}
\end{equation}
}

\noindent where the update functions $S_n^\text{img}$, $S_n^\text{sys}$, $S_n^\text{actor}$, $S_n^\text{critic}$ and $S_n^\text{dec}$ are stochastic gradients with loss functions (to be described in following sections) calculated with data samples from the replay buffers, i.e., $\mathcal{M}_{\text{trajectory}}$, $\mathcal{M}_{\text{transition}}$, and $\mathcal{M}_{\text{decision}}$. The replay buffers store a finite number of recently observed tuples of $(\vect{x}_t, \vect{a}_t)$, $(\vect{h}_{t-1}, \vect{a}_t, r_t, \vect{h}_t)$, and $(\vect{h}_t, \vect{a}_t, l_t)$ in $\mathcal{M}_{\text{trajectory}}$, $\mathcal{M}_{\text{transition}}$, and $\mathcal{M}_{\text{decision}}$ respectively.

The step sizes for the various components of our multi-time scale optimization, namely $\epsilon_n^\text{img}$, $\epsilon_n^\text{dec}$, $\epsilon_n^\text{actor}$, $\epsilon_n^\text{critic}$, and $\epsilon_n^\text{sys}$, are determined as follows. The generation of an adversarial image perturbation depends on the generator with parameter $\theta_n^\text{img}$,
the actor that determines the attack coordinates with parameter $\theta_{n}^\text{actor}$, and the binary decision maker that chooses whether to use the adversarial perturbation or not with parameter $\theta_{n}^\text{dec}$. As the generation of the adversarial image perturbation and its use are governed by a policy with parameters $\theta_{n}^\text{img}$, $\theta_{n}^\text{actor}$, and $\theta_{n}^\text{dec}$, we set faster update rates for the parameters that are relevant to policy evaluation, i.e., $\theta_{n}^\text{critic}$ and $\theta_{n}^\text{sys}$.
Hence, the step size follows the diminishing rules as   $n \rightarrow \infty$
{\small
\begin{equation}\label{eq:step-size-rule}
    \frac{\epsilon_n^\text{img}}{\epsilon_n^\text{dec}} \rightarrow 0 \quad
    \frac{\epsilon_n^\text{dec}}{\epsilon_n^\text{actor}} \rightarrow 0 \quad
    \frac{\epsilon_n^\text{actor}}{\epsilon_n^\text{critic}} \rightarrow 0 \quad
    \frac{\epsilon_n^\text{critic}}{\epsilon_n^\text{sys}} \rightarrow 0,
\end{equation}
}

\noindent This is because we intend to set slower update rates for the lower-level components of the policy that generate data for the upper-level components of policy evaluation.
The multi-time scale stochastic optimization is summarized in Algorithm~\ref{alg:train_attacker} in the appendix of the extended version~\cite{yoon2021learning}.

We describe the loss functions of the stochastic gradients for the multi-level stochastic optimization as follows:

\subsubsection{Image attack generator}\label{sec:image_attack_generator}
We utilize a \emph{white box} model as a proxy object detector to train the attack generator. Specifically, we use the recently released version of \emph{YOLO}, called \emph{YOLOv5}~\cite{yolov5}, for this purpose.

\begin{figure}[h]
     \centering
     \begin{subfigure}[b]{0.32\columnwidth}
         \centering
         \includegraphics[width=\columnwidth]{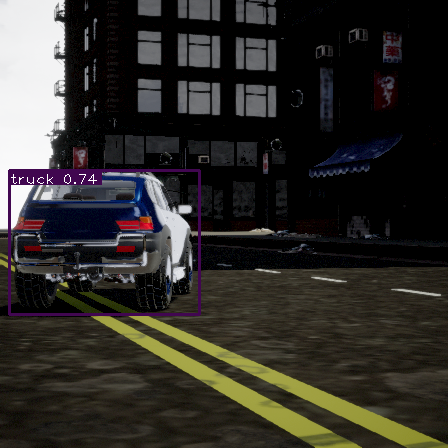}
         \caption{image}
     \end{subfigure}
     \hfill
     \begin{subfigure}[b]{0.32\columnwidth}
         \centering
         \includegraphics[width=\columnwidth]{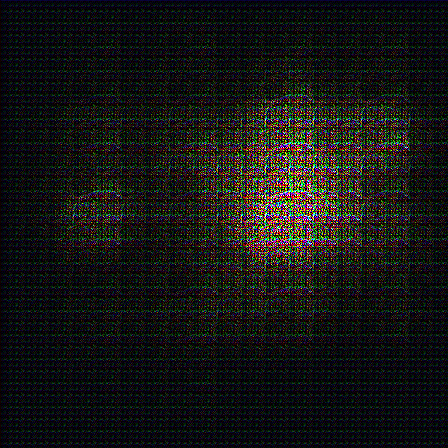}
         \caption{perturbation $\vect{w}_t$}
     \end{subfigure}
     \hfill
     \begin{subfigure}[b]{0.32\columnwidth}
         \centering
         \includegraphics[width=\columnwidth]{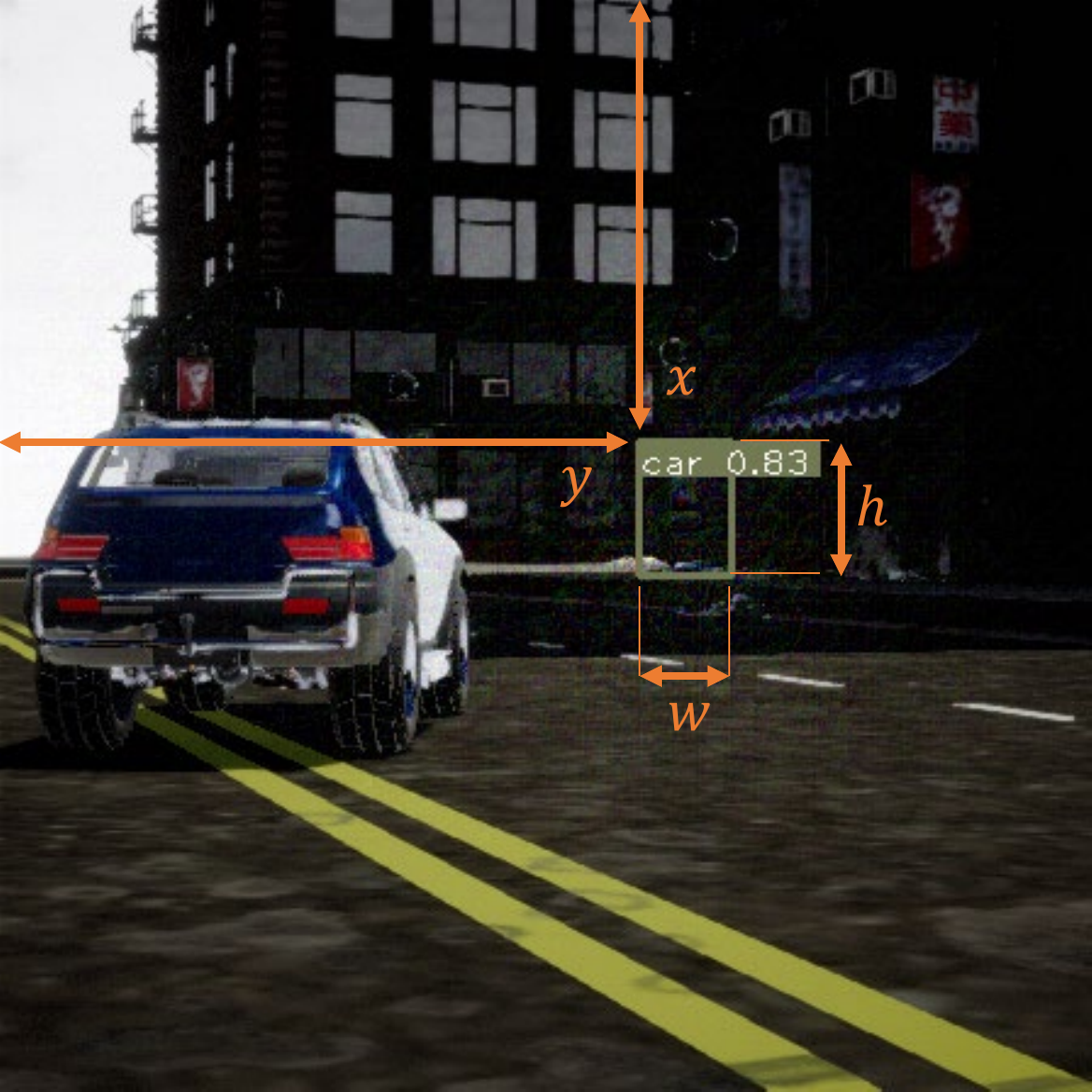}
         \caption{perturbed image}
     \end{subfigure}
        \caption{Fabrication of the bounding box at $(x, y, w, h)$ with $\vect{w}_t$}
        \label{fig:one_time_attack_synthesis}
\vskip -0.1in
\end{figure}

Our image attack generator fabricates bounding boxes at the target coordinates, injecting adversarial perturbations as shown in Figure~\ref{fig0_adv_img_attack}b. Using reinforcement learning, the high-level attacker (reinforcement learning agent) selects the target coordinates to place the fabricated bounding boxes accordingly. Given the high-level attack $\mathbf{a}_t \in [0,1]\times[0,1]\times[0,1]\times[0,1]$ representing the coordinates $x$ and $y$ of a bounding box, its width and height, the image attack network aims to delete all other bounding boxes but keep the one corresponding to the high-level attack as illustrated in Figure~\ref{fig:one_time_attack_synthesis}. By performing optimization iterations (500 iterations in Figure~\ref{fig:one_time_attack_synthesis}), we can delete the existing bounding box and place a bounding box according to the target coordinates.

The iterative optimization approach that creates a bounding box for online image attacks, as shown in Figure~\ref{fig:one_time_attack_synthesis}, is not suitable as it must be performed within a fixed time step of the control loop in the autonomous system. In our framework, we instead train an attack generator that minimizes the loss function through the image attack generator:
{\small
\begin{equation}\label{eq:argmin_img_attack_loss}
\argmin_{\theta^\text{img}} l^{img}\left(\vect{w}(\vect{x};\theta^\text{img}), \vect{x}, \vect{a} \right),
\end{equation}
}

\noindent where $\vect{w}(\vect{x};\theta^\text{img})) := \textbf{Dec}_1\left(\vect{Enc}_1(\mathbf{x}; \theta^\text{img}), \vect{a}; \theta^\text{img}\right)$, and $\vect{x}$ and $\vect{a}$ are sampled from $\mathcal{M}_\text{trajectory}$. This approach differs from the optimization over image space that is suitable for one-time use, i.e., $\argmin_{\vect{w}} l^{img}(\vect{w}, \vect{x}, \vect{a})$. The stochastic gradient $S_n^\text{img}(\mathcal{M}_\text{trajectory})$ in~\eqref{eq:multi-time-scale} is associated with the loss function in~\eqref{eq:argmin_img_attack_loss}. To fabricate the detected bounding box, i.e., inverted mapping from the fabricated detection to the input image,
we use the loss function employed from~\cite{redmon2016you}, where the same loss function is used to train the YOLO detector. The specific loss function in~\eqref{eq:argmin_img_attack_loss} from~\cite{redmon2016you} is described in the appendix of the extended version~\cite{yoon2021learning}.

\subsubsection{System identification for state estimation}\label{sec:state_estimation_learning}
Due to incomplete observations of the state $\vect{s}_t$ through image stream $\vect{x}_t$, we need to identify the system to construct the estimator. The system identification determines the parameter that maximizes the state estimate's likelihood. We maximize the likelihood of state predictor by minimizing the cross-entropy error between true image streams and the predicted image streams by a stochastic optimization which samples trajectories saved in the memory buffer denoted by $\mathcal{M}_\text{trajectory}$ with a loss function to minimize.

The loss function $l^\text{sys}(\cdot)$ is calculated using the sampled trajectories from $\mathcal{M}_{\text{trajectory}}$
We calculate the loss function as
{\small
\begin{equation}\label{eq:loss_sys_id}
    l^\text{sys}(\mathcal{M}_{\text{trajectory}};\theta_\text{sys}) = \frac{1}{M}\sum_{m=1}^M H(\mathbf{X}_m, \hat{\mathbf{X}}_m)
\end{equation}
}
\noindent where $\mathbf{X}_m = (\mathbf{x}_0, \dots, \mathbf{x}_T)_m$ is the $m$\textsuperscript{th} sample image stream with time length $T$. Here, $H(\cdot, \cdot)$ is average of the binary cross-entropy $h(\cdot, \cdot)$ between the original image stream $\mathbf{X}_m$ and the predicted image stream $\hat{\mathbf{X}}_m$, which is computed over the RGB pixel values of the image streams.

We generate the predicted trajectory, $\hat{\mathbf{X}}_m =\{\hat{\mathbf{x}}_1, \dots, \hat{\mathbf{x}}_T\}$, given the original trajectory with image stream $\mathbf{X}_m = (\mathbf{x}_0, \dots, \mathbf{x}_T)_m$ and action stream $(\mathbf{a}_0, \dots, \mathbf{a}_T)_m$ by processing them through the encoder, GRU, and the decoder as
{\small
\begin{equation*}
    \begin{aligned}
    \mathbf{h}_{t+1} &= \text{GRU}(\mathbf{h}_t, \text{Encoder}_1(\mathbf{x}_t), \mathbf{a}_t), \quad \mathbf{h}_0 \sim \mathcal{N}(0,\mathbf{I}),\\
    \hat{\mathbf{x}}_{t+1} & = \text{Decoder}_1(\mathbf{h}_{t+1}).
    \end{aligned}
\end{equation*}
}

\noindent With the loss function in~\eqref{eq:loss_sys_id}, the stochastic gradient for the optimization is defined as $S^\text{sys}_n =  -\nabla_{\theta_\text{sys}} l^\text{sys}(\mathcal{M}_\text{trajectory};\theta_\text{sys})$.

\subsubsection{Actor-Critic policy improvement}
The attack coordinate $\vect{a}_t$ is determined by the policy, i.e., $\vect{a}_t=\mu(\mathbf{h}_t; \theta_\text{actor})$ that maps the state estimate $\mathbf{h}_t$ into an action $\vect{a}_t$. To improve the policy, the critic evaluates the policy relying on the principle of optimality~\cite{bellman1966dynamic}. We employed an actor-critic method~\cite{lillicrap2015continuous} for the reinforcement learning agent in the proposed framework. The critic network is updated using the state estimate $\vect{h}_t$ to apply the optimality principle with the following stochastic gradient as $S^\text{critic}_n = -\nabla_{\theta_\text{critic}} l^\text{critic}(\mathcal{M}_\text{transition};\theta_\text{critic})$
with the following loss function
{\small
\begin{equation*}
\begin{aligned}
    l^\text{critic}(\mathcal{M}_{\text{transition}};\theta^\text{critic})=\frac{1}{M}\sum_{m=1}^M(Q(\mathbf{h}_m, \mathbf{a}_m; \theta^\text{critic})- Q_m^{\text{target}})^2,
\end{aligned}
\end{equation*}
}

\noindent where $Q_m^{\text{target}} = r_m + \gamma Q(\mathbf{h}'_{m}, \mu_\theta(\mathbf{h}'_m);\theta_\text{critic})$ and the state transition samples, i.e., $\left((\mathbf{h}, \mathbf{a}, \mathbf{h}', r)_0, \dots, (\mathbf{h}, \mathbf{a}, \mathbf{h}', r)_M\right)$, are sampled from the replay buffer $\mathcal{M}_{\text{transition}}$.
With the same state transition data samples, we calculate the stochastic gradient for the policy update as $S^\text{actor}_n = \nabla_{\theta_\text{actor}} J(\mathcal{M}_\text{transition};\theta_\text{actor})$ with the following estimated value function as
{\small
\begin{equation*}
\begin{aligned}
    J = \frac{1}{M}\sum_{m=1}^M Q(\mathbf{h}_m, \mu(\mathbf{h}_m; \theta_\text{actor});\theta_\text{critic}),
\end{aligned}
\end{equation*}
}
where $Q(\mathbf{h}, \vect{a};\theta_\text{critic})$ indicates the value of taking action $\vect{a}$ at the state estimated as $\vect{h}$.
\subsubsection{Loss estimators for the binary decision making}
The stochastic gradient $S^\text{dec}_n$ in~\eqref{eq:multi-time-scale} is associated with the two optimizations described in~\eqref{eqn:attack_loss_prediction}. The entire stochastic optimization with the aforementioned stochastic gradients is summarized as Algorithm~\ref{alg:train_attacker} in the appendix of the extended version~\cite{yoon2021learning}.

\section{Experiments}\label{sec:experiment}
\begin{figure}[h]
     \centering
     \begin{subfigure}[b]{0.45\columnwidth}
         \centering
         \includegraphics[width=\columnwidth]{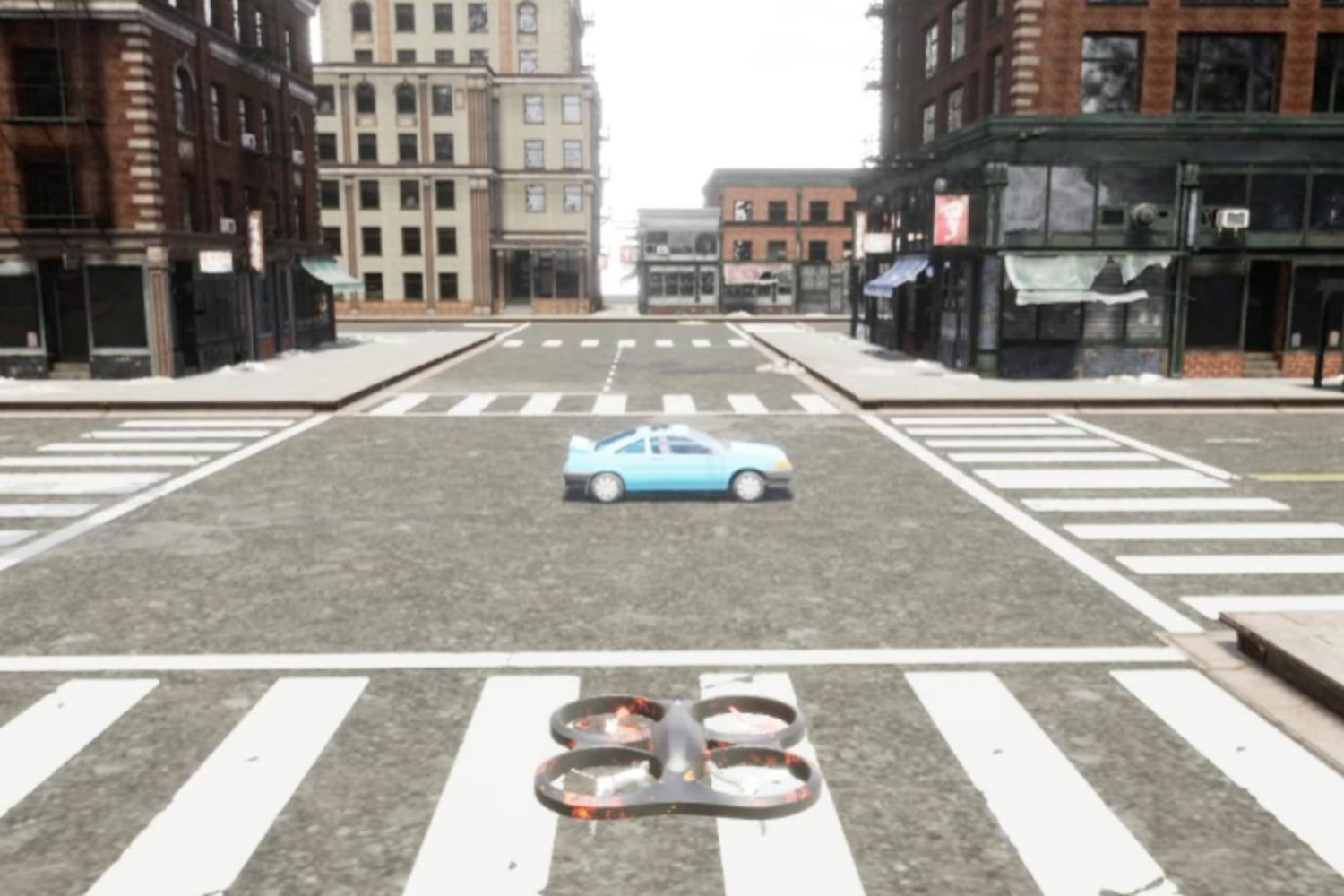}
         \caption{Drone to a car}
     \end{subfigure}
     \hfill
     \begin{subfigure}[b]{0.45\columnwidth}
         \centering
         \includegraphics[width=\columnwidth]{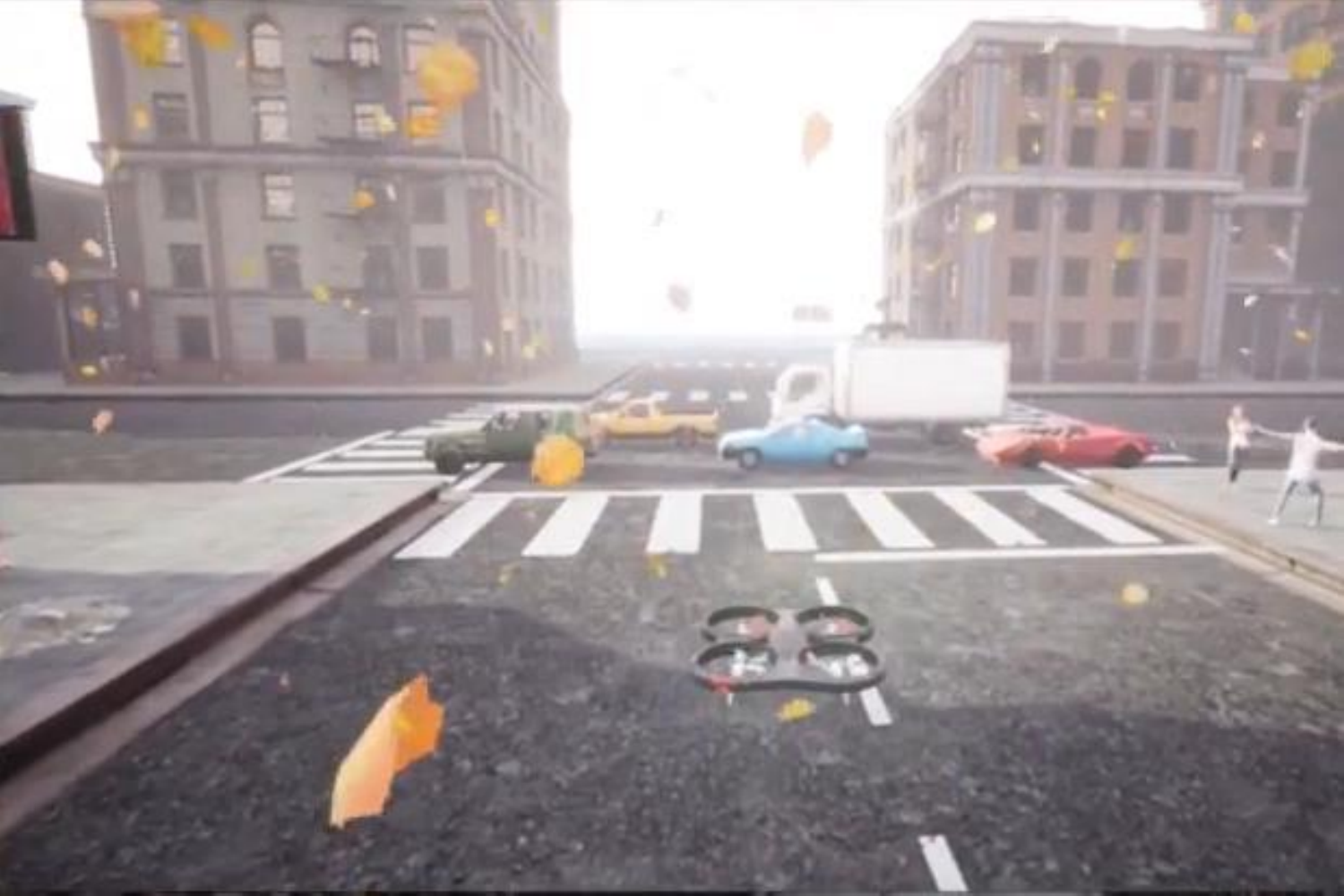}
         \caption{Cars and trucks}
     \end{subfigure}
     \hfill
     \begin{subfigure}[b]{0.45\columnwidth}
         \centering
         \includegraphics[width=\columnwidth]{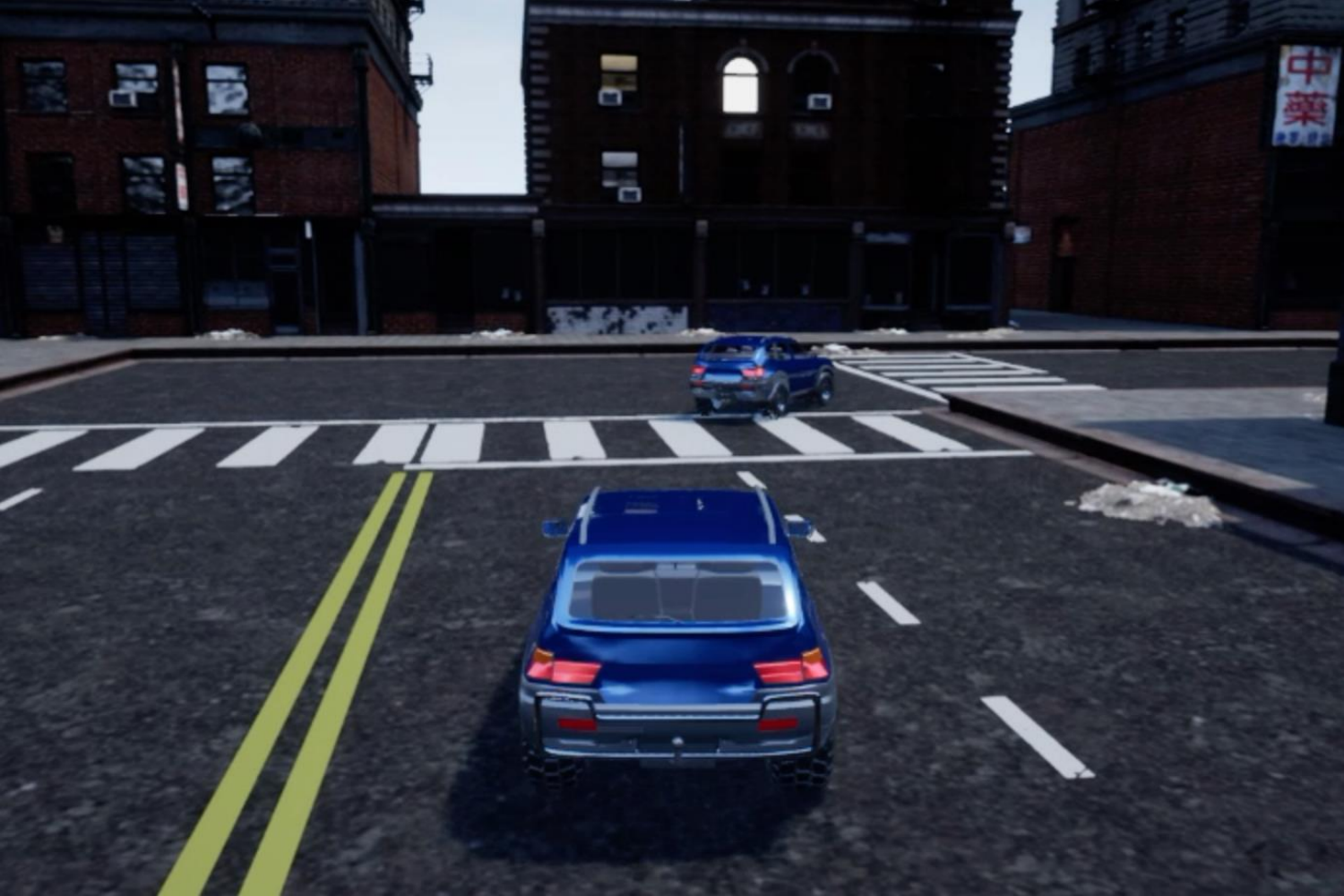}
         \caption{Following a car}
     \end{subfigure}
     \hfill
     \begin{subfigure}[b]{0.45\columnwidth}
         \centering
         \includegraphics[width=\columnwidth]{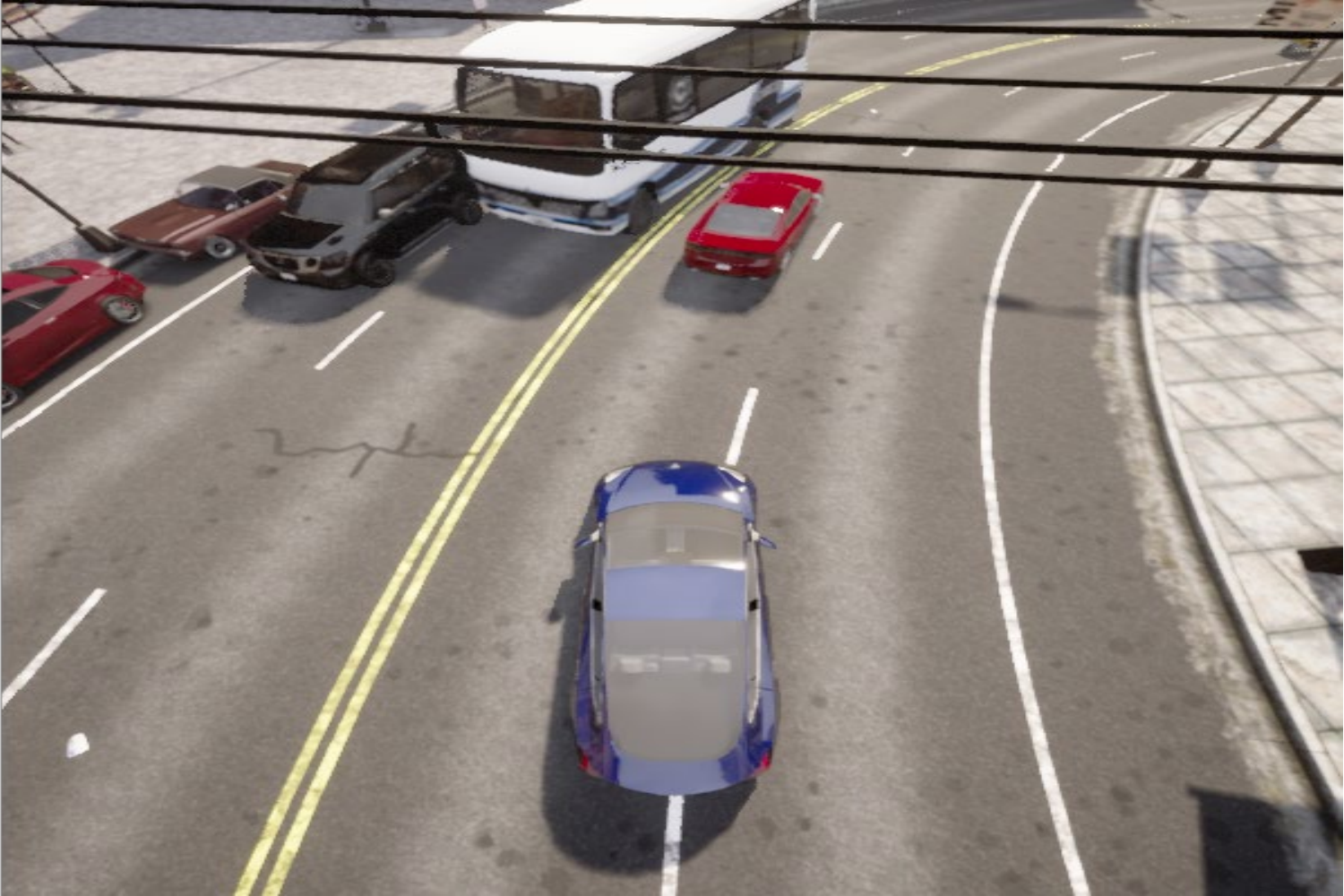}
         \caption{Following a car in \underline{traffic}}
     \end{subfigure}
     \caption{Simulation environments.}
        \label{fig:environments}
\vskip -0.2in
\end{figure}

We tested the proposed attack method to determine its ability to mislead autonomous vehicles in line with adversarial objectives. The adversary relied solely on image frames as sensing input and an uncertain actuator, in the form of an adversarial perturbation, to manipulate the paths of autonomous vehicles. Despite the adversary's limited sensor and uncertain actuator, our proposed algorithm successfully misled the autonomous vehicles in various simulation environments shown in Figure~\ref{fig:environments} (and in illustrative videos\footnote{(a) \emph{Drone to a car} at \href{https://youtu.be/sjgQGgyLR8Y}{link1}; (b) \emph{Cars and trucks} at \href{https://youtu.be/Xx9hH6mP0PE}{link2}; (c) \emph{Following a car} at \href{https://youtu.be/mOPfPDEXkdM}{link3}; (d) \emph{Following a car in traffic} at \href{https://youtu.be/z61FyoJx_Yg}{link4}}).

\begin{figure}[ht]
  \begin{center}
    \includegraphics[width=\linewidth]{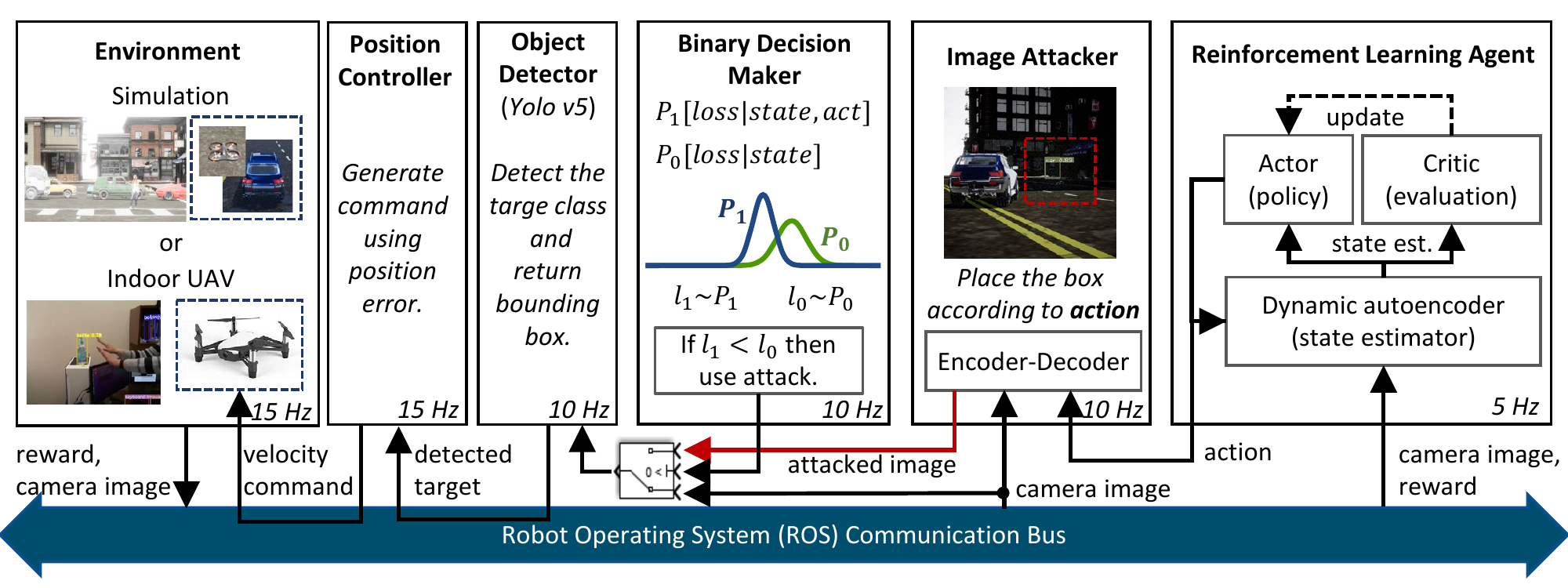}
  \end{center}
  \caption{The framework implemented using \emph{ROS}.}
  \label{fig_ros_implementation}
\vskip -0.2in
\end{figure}

All our experiments consider attacking a vision-based guidance system depicted in Figure~\ref{fig1_victim_sys_malware} that uses \emph{YOLOv5} object detector~\cite{yolov5}. To simulate the vehicle environment, we employed a game development editor (\emph{Unreal}~\cite{Unreal}) that is capable of building \emph{photo-realistic} 3D environments, along with plug-in tools such as \emph{AirSim}~\cite{airsim2017fsr} and \emph{CARLA}~\cite{Dosovitskiy17}. For the attack algorithm implementation, we used the robot operating system (\emph{ROS}) to simultaneously implement the attack model learning and executing the attack using multiple modules (nodes) as illustrated in Figure~\ref{fig_ros_implementation}. All experiments were conducted on a desktop computer equipped with a GPU capable of rendering the 3D environments and performing DNN training.

\subsection{Baseline method}
\begin{figure}[ht]
\vskip -.05in
\begin{center}
\centerline{\includegraphics[width=0.7\columnwidth]{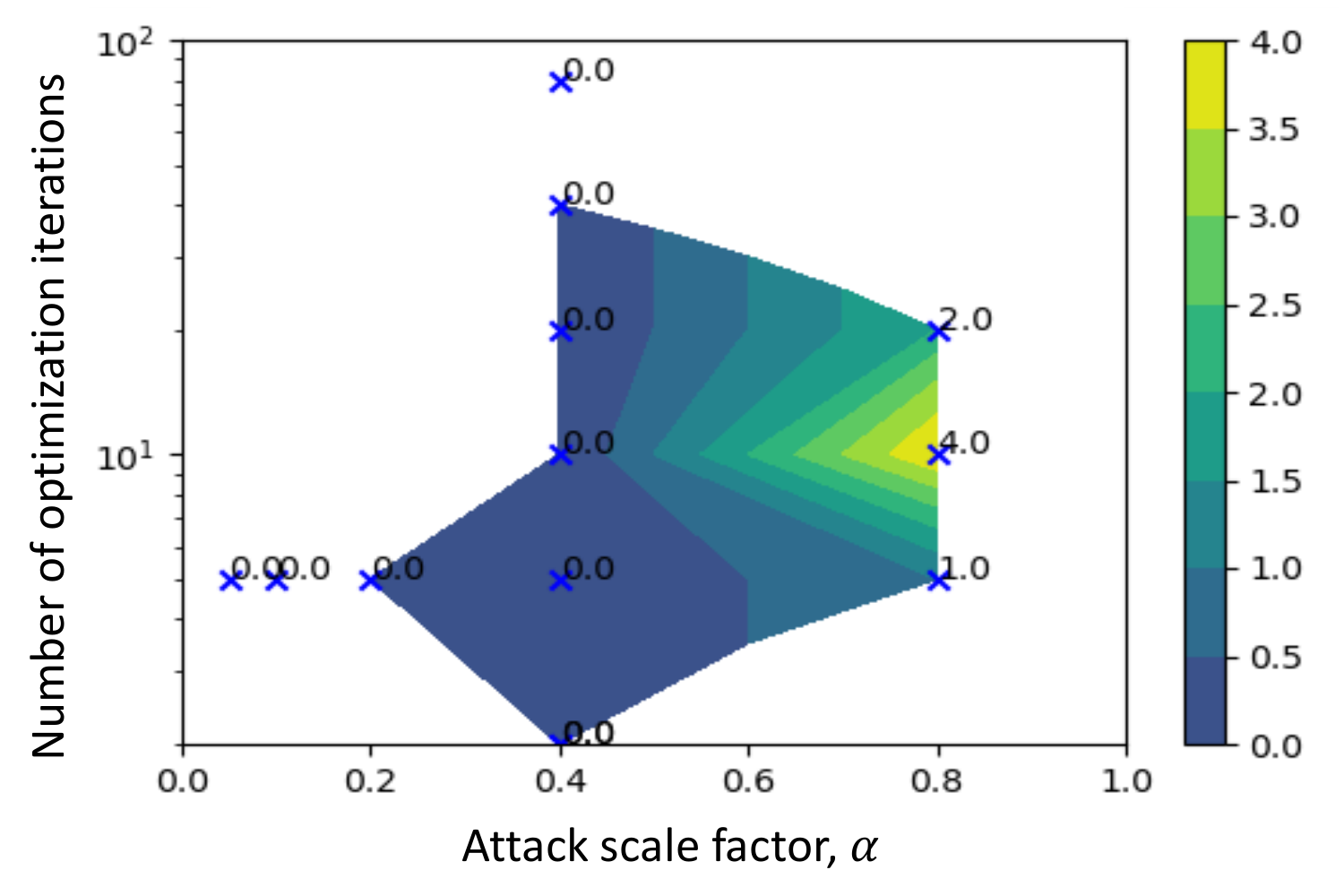}}
\caption{Terminal rewards in the 3\textsuperscript{rd} environment (Figure~\ref{fig:environments}c) of the baseline method with different of $\alpha$ and the iterations.}
\label{fig_baseline_param_sweeping}
\end{center}
\vskip -.2in
\end{figure}

Our proposed framework was compared to an image attack method presented in~\cite{jia2020fooling, jha2020ml}, which uses iterative optimization with the image tensor as the decision variable. The effectiveness of these methods depends on the number of iterations and the scale factor $\alpha$, which limits the size of the adversarial perturbation as $\tilde{\mathbf{x}}_t= \max(\min(\mathbf{x}_t + \alpha \mathbf{w}_t, 1),0)$. Previous works~\cite{jia2020fooling, jha2020ml} developed such methods as offline approaches. When an infinite number of iterations are allowed, the offline method can arbitrarily fabricate the bounding box, as illustrated in Figure~\ref{fig:one_time_attack_synthesis}. To the best of our knowledge, no online image attack methods have been applied to autonomous vehicles. Therefore, we set a baseline method by applying the iterative optimization method as an online algorithm. In addition to the number of iterations, the scale factor $\alpha$ is a critical hyperparameter, as a higher value can increase the attacker's ability to fabricate bounding boxes, but it also reduces the stealthiness of the attack. For example, we collect the performance of the base line method with varying number of iteration and the scale factor as shown in Figure~\ref{fig_baseline_param_sweeping}.
For the \underline{first three experiments} (the first three rows in Table~\ref{tb:comparative_analysis}), we set the \underline{hyperparameter $\alpha=0.8$} and 20 iterations for the baseline method (Iterative optimization). The optimization method generated image perturbations approximately every 1 second. However, for our proposed methods, including Generative Attack, Recursive Attack, Neural Thompson, and Conditional Sampling, we set $\alpha=0.05$ for stealthiness of the image attack. Additionally, the baseline method required manual annotation of the bounding box to be fabricated, as described in~\cite{open_review}. Therefore, we manually placed the target area to fabricate the bounding box according to the adversarial objectives, such as placing the bounding box to the left when the adversary needs to move the vehicle to the left. In the \underline{last experiment}, we set the scale factor of the baseline method \underline{equal to that of our proposed methods}, i.e., $\alpha=0.05$. For $\alpha$ values greater than 0.2, the baseline method was as effective as our proposed methods.
    
\subsection{Ablation study}
We evaluated the proposed framework by conducting ablations, as presented in Table~\ref{tab:methods}. To test the effectiveness of each method, we conducted experiments in four different environments illustrated in Figure~\ref{fig:environments}. In the 1\textsuperscript{st} environment (Figure~\ref{fig:environments}a), we set the reward as the distance between the host vehicle and the target object. Thus, the adversary can increase the distance to move the vehicle away from the target. In the 2\textsuperscript{nd} environment (Figure~\ref{fig:environments}b), the reward is set as the horizontal coordinate of the host vehicle with respect to the target object. In this scenario, learning to increase the horizontal coordinate would lead to a crash. In the 3\textsuperscript{rd} environment (Figure~\ref{fig:environments}c), the adversary learns to increase the distance from the front car. In the fourth environment, we rewarded the learning agents when the distance between the host vehicle and the front car was greater than 50 meters. As shown in Table~\ref{tb:comparative_analysis}, the four environments have different object tracking methods. The first two environments use only \textbf{YOLO} detection. In the 3\textsuperscript{rd} environment, a Kalman filter is used to filter out the changes of the bounding boxes that is the outcome of the detecion. In the last environment, an multi-object tracking (SORT~\cite{bewley2016simple}) is implemented to deal with multiple cars in the traffic. We report the performance of the trained attackers (listed in Table~\ref{tab:methods}) with the last ten episodes of the entire 200 training episodes in Table~\ref{tb:comparative_analysis}.
\begin{table}[h]
    \centering
    \begin{tabular}{l|ccc}
\hline
                      &Generative &State     &Attack  \\
Methods               &network    &estimator &switch  \\
                      &(Y/N)      &(Y/N)     &(Y/N)   \\
\hline
Iterative optimization& N         & N        &N  \\
 Generative attack    & Y         & N        &N  \\
 Recursive attack     & Y         & Y        &N  \\
 Neural Thompson sampling    & Y         & Y        &Y  \\
 Conditional sampling & Y         & Y        &Y  \\
\hline
\end{tabular}
    \caption{List of components for ablation study.}
    \label{tab:methods}
\vskip -0.1in
\end{table}

In comparison to the baseline iterative optimization method, the recursive attack methods demonstrated higher terminal rewards, collision rates, and terminal distances. The incorporation of the state estimator improved the overall performance in terms of terminal rewards. Since we desire infrequent use of the image attack for stealthiness, we measure how frequently the attacks are used, i.e., attack rate. The utilization of binary decision-makers such as Neural Thompson Sampling (NTS) or the proposed conditional sampling resulted in decreased attack rates and the difference between the attacked images and the original images in terms of L2 norm and SSIM loss. Moreover, our proposed conditional sampling method showed higher terminal rewards compared to NTS. 

\begin{table*}[t]
\vskip -0.2in
\centering
\begin{tabular}{c|c|l|ccc|ccc}
\hline
\multirow{3}{*}{Environments}    &\multirow{3}{*}{Object Tracking}&                       & Attack    &SSIM      &L2         & Collision   &Terminal  &Time AVG\\
                                 &     &Methods                & rate      &loss      &loss       & rate        &reward    &reward\\
                                 &     & &(\%)&($10^{-2}$, avg)&($10^{-4}$, avg)                  &(\%) &(avg$\pm$stdev)   &(avg$\pm$stdev)\\
\hline
                                 &                &Normal                 &-          & -        & -         &0                &$15.7\pm0.0$ &$1.29\pm0.07$\\
                                 &                &Iterative optimization &$100$      & $13.6$   & $12.1$    &0                &$15.7\pm0.1$ &$1.29\pm0.07$\\
        Drone to a car           &YOLO           &Generative attack      &$100$      & $8.74$   & $3.94$    &74               &$\mathbf{ 30.1\pm8.3}$ &$\mathbf{2.58\pm0.80}$\\
(Figure~\ref{fig:environments}a)    &detection only &Recursive attack       &$100$      & $17.4$   & $7.61$    &$\mathbf{76}$    &$29.7\pm8.1$ &$2.16\pm0.45$\\
                                 &&Neural Thompson        &$\mathbf{31.7}$& $\mathbf{6.81}$&$\mathbf{2.58}$&0  &$21.5\pm10.5$&$1.64\pm0.49$\\
                                 &&Conditional sampling   &$55.3$     & $9.15$   & $4.11$    &52               &$27.8\pm9.2$ &$2.16\pm0.57$\\
\hline
                                 &&Normal                 &-          & -        & -         &0                &$1.8\pm1.8$  &$0.10\pm0.15$\\
                                 &&Iterative optimization &$100$      & $16.7$   & $30.3$    &0                &$-0.4\pm2.3$ &$-0.03\pm0.13$\\
    Cars and trucks              &YOLO &Generative attack      &$100$      & $7.52$   & $2.68$    &18               &$5.4\pm13.3$ &$0.29\pm0.75$\\
  (Figure~\ref{fig:environments}b)  &detection only &Recursive attack       &$100$      & $7.26$   & $2.74$    &$\mathbf{36}$    &$\mathbf{9.4\pm6.3}$  &$\mathbf{0.87\pm0.43}$\\
                                 &&Neural Thompson        &$49.3$     & $4.94$   & $1.77$    &8                &$6.0\pm9.2$  &$0.25\pm0.46$\\
                                 &&Conditional sampling   &$\mathbf{42.7}$     & $\mathbf{1.97}$   & $\mathbf{0.81}$    &20   &$2.6\pm10.6$ &$0.10\pm0.53$\\
\hline
                                 &&Normal                 &-          & -        & -        &$0$              &$0\pm0$     &$4.25\pm0.04$\\
                                 &YOLO&Iterative optimization &$100$        & $9.17$   & $7.58$  &$30$              &$3.0\pm4.6$ &$4.17\pm0.23$\\
    Following a car              &detection& Generative attack      &$100$        & $5.96$   & $3.16$  &$66$              &$7.0\pm4.4$ &$2.92\pm1.10$\\
    (Figure~\ref{fig:environments}c)&and&Recursive attack       &$100$        & $4.88$   & $2.61$  &$74$               &$7.4\pm4.4$ &$2.92\pm1.33$\\
                    &\underline{Kalman filter} &Neural Thompson        &$\mathbf{33.5}$&$\mathbf{1.66}$   &$\mathbf{0.96}$&$52$              &$5.2\pm5.0$ &$4.01\pm0.39$\\
                                 &&Conditional sampling   &$72.8$     & $3.10$   & $2.06$    &$\mathbf{76}$              &$\mathbf{7.6\pm4.3}$ &$\mathbf{4.27\pm3.54}$\\
\hline
                                 &YOLO&Normal                 &-        & -        & -         &$6$              &$3.7\pm4.2$     &$\mathbf{2.28}\pm0.30$\\
                                 &detection &Iterative optimization &$100$        & $0.0$        & $0.0$         & $4$              &$3.2\pm3.5$     &$2.24\pm0.34$\\
    Following a car              &and&Generative attack      &$100$        & $20.1$        &$9.1$         & $\mathbf{30}$              &$\mathbf{10.4}\pm3.7$     &$0.97\pm0.72$\\
    in \underline{traffic} (Figure~\ref{fig:environments}d)       &\underline{multi-object}&Recursive attack       &$100$        &$23.5$        &$10.6$         & $2$  &$9.9\pm2.2$     &$0.63\pm0.60$\\
                                 &\underline{tracking}      &Neural Thompson        &$\mathbf{34.6}$        &$\mathbf{8.1}$        &$\mathbf{3.7}$         &$10$              &$8.1\pm5.3$     &$1.39\pm0.83$\\
                                 &(SORT)                    &Conditional Sampling   &$49.6$        &$12.4$        &$5.56$         &$10$              &$9.3\pm4.8$     &$1.09 \pm 0.79$\\
\hline
\end{tabular}
\caption{\small Ablation study with the last 10 episodes in 5 random training experiments, i.e., $N=50$. The stealthiness is evaluated by Attack rate, SSIM loos, and L2 loss. And the attacker's performance to disrupt is evaluated by Collision rate, Terminal reward, and Time averaged reward.}
\label{tb:comparative_analysis}
\vskip -0.2in
\end{table*}

\begin{figure}[ht]
\vskip -.1in
\begin{center}
\centerline{\includegraphics[width=0.7\columnwidth]{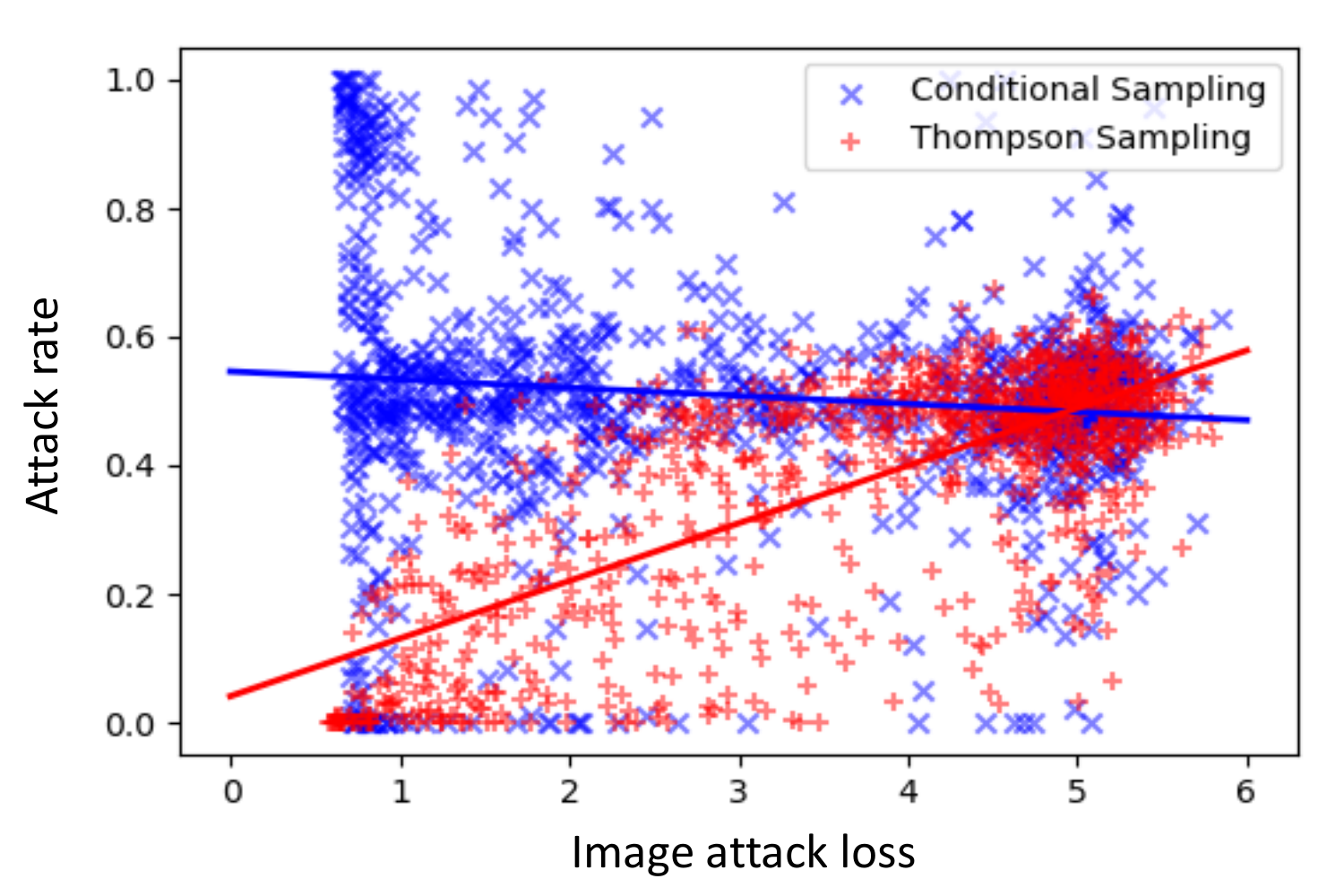}}
\caption{Attack rate vs. Image attack loss of the 5 training experiments with the 2\textsuperscript{nd} environment (Figure~\ref{fig:environments}b).}
\label{fig_correlation}
\end{center}
\vskip -.4in
\end{figure}

Moreover, the conditional sampling shows the higher use of the attack when the image attack loss is lower as we intended as shown in Figure~\ref{fig_correlation}. In contrast, the Thompson sampling (NTS) shows the opposite correlation, i.e., using the image attack when the loss values are higher. Further information regarding the simulations can be found in the appendix  of the extended version~\cite{yoon2021learning}. 

\subsection{Real robot experiment}\label{sec:real_robot_experiment}
\begin{figure}[ht]
 \vskip -0.1in
  \begin{center}
    \includegraphics[width=1.0\linewidth]{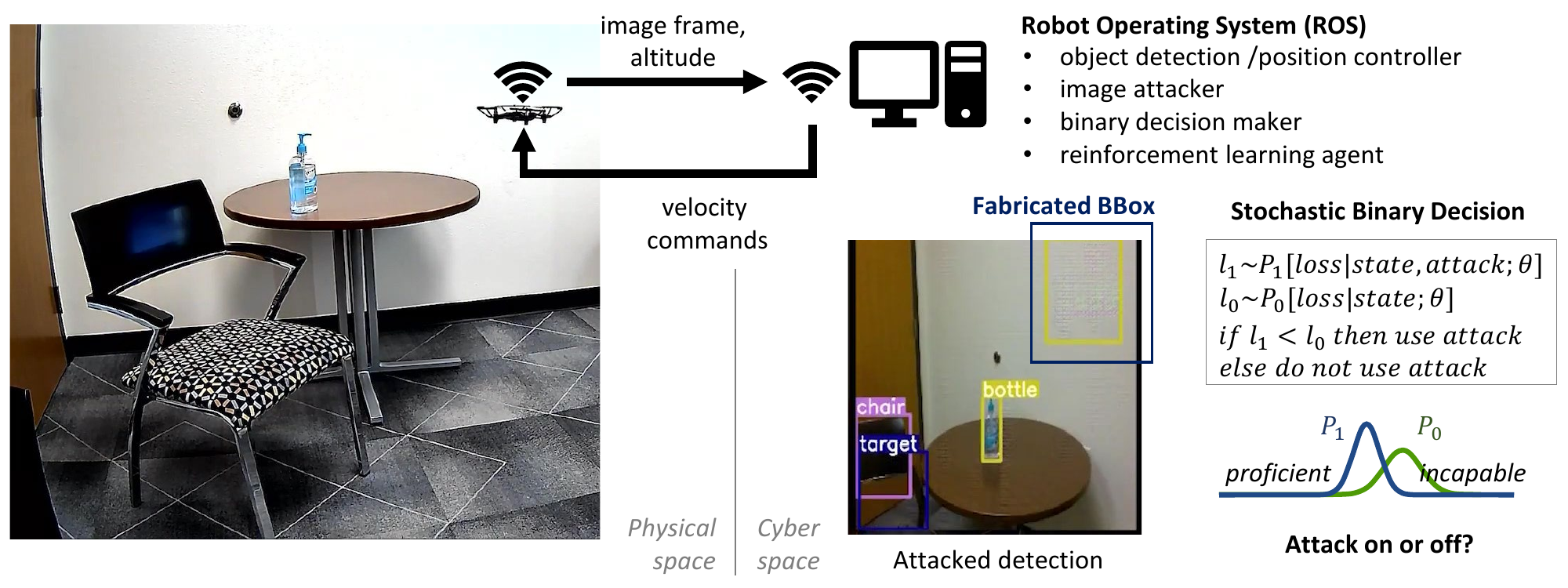}
  \end{center}
  \caption{Attacking UAV's visual-tracking of a bottle.}
  \label{fig_indoor_robot}
\vskip -0.1in
\end{figure}

We implemented the proposed framework with a miniature drone (\emph{DJI Tello}) to validate its efficacy in real-world scenarios, as depicted in Figure~\ref{fig_ros_implementation}. The drone employs IMU, optical-flow, and barometer for velocity estimation and to follow velocity commands. The drone was connected to a desktop computer via wifi-networks, as shown in Figure~\ref{fig_indoor_robot}. The objective of the online image training was to crash the UAV by teaching it to fabricate the bounding box.
In the linked video\footnote{Illustrative video available at \href{https://youtu.be/4w0pvQRCVHc}{https://youtu.be/4w0pvQRCVHc}}, the online training lasted for ten minutes, and the UAV crashed successfully.

\section{Conclusion}
This work showed a new online image attack framework that improves the iterative optimization-based methods that are more suitable for offline attack generation. In our proposed framework, the image attacks can be generated in real-time using only the image stream collected from the autonomous vehicle. Furthermore, the proposed conditional sampling for the binary decision making whether to use the attack (or not) improves the stealthiness by waiting until the proficiency increases. This work will serve as a stepping stone towards strengthening the perception in autonomous vehicles by learning worst-case attack scenarios.

\bibliographystyle{IEEEtran}
\bibliography{mybib}




\appendix
\subsection{Image attack loss function.}
 The loss function of the adversarial perturbation $\vect{w}$ given an image $\vect{x}$ and an action $\vect{a}$ that describe the target bounding box coordinates $(x, y, w, h)$ is as follows:
{\small
\begin{equation}\label{eq:loss_img_attack}
\begin{aligned}
     &l^\text{adv}(\mathbf{w}; \mathbf{x}, \mathbf{a})\\
     &= \lambda_{\text{coord}} \sum_{i=1}^{3}\sum_{j=1}^{S_i^2}\sum_{k=1}^{B} \mathbf{1}_{i,j,k}^\text{obj}\left[(x -\hat{x}_{i,j,k})^2 + (y -\hat{y}_{i,j,k})^2\right] \\
     &+ \lambda_{\text{coord}} \sum_{i=1}^{3}\sum_{j=1}^{S_i^2}\sum_{k=1}^{B} \mathbf{1}_{i,j,k}^\text{obj}\left[(\sqrt{w} -\sqrt{\hat{w}_{i,j,k}})^2 + (\sqrt{h} -\sqrt{\hat{h}_{i,j,k}})^2\right]\\
     &+ \sum_{i=1}^{3}\sum_{j=1}^{S_i^2}\sum_{k=1}^{B} \mathbf{1}_{i,j,k}^\text{obj}(1 - \hat{C}_{i,j,k})^2 \\
     &+ \lambda_{\text{no obj}} \sum_{i=1}^{3}\sum_{j=1}^{S_i^2}\sum_{k=1}^{B} \mathbf{1}_{i,j,k}^\text{no obj}(0 - \hat{C}_{i,j,k})^2 \\
     &+ \sum_{i=1}^{3}\sum_{j=1}^{S_i^2}\sum_{k=1}^{B}\mathbf{1}_{i,j}^\text{no obj}\sum_{c\in\text{classes}}(p_{i,j,k}(c) - \hat{p}_{i,j,k}(c))^2
\end{aligned}
\end{equation}
}

\noindent where $\mathbf{1}_{i,j,k}^\text{obj}=1$ if the target object is associated with the grid of the index $i,j,k$, and otherwise, it is zero. Also, $\mathbf{1}_{i,j,k}^\text{no obj}=1$ if $\mathbf{1}_{i,j,k}^\text{obj}=0$, and otherwise it is zero. In~\eqref{eq:loss_img_attack}, the first two terms are to minimize the error between inferred coordinates of the bounding box and the target coordinates $x ,y, w, h$ that are mapped from $\mathbf{a}_t$ through linear equations. The first two terms are added into the loss function with the weight $\lambda_{\text{coord}}$.

In~\eqref{eq:loss_img_attack}, the third and the fourth term count the error in having a target object associated with each anchor box with index $i, j, k$. The fourth term is added to the loss function with the weight $\lambda_{\text{no obj}}$. The values of the indicator $\mathbf{1}_{i,j,k}^\text{obj}$ given the target coordinate $(x,y,w,h)$ is determined using the following procedure. 

First, we select $(i', k')$ through the minimization below
{\small
\begin{equation*}
    i', k' = \argmin_{i,k} \left(\sqrt{w^\text{template}_{i,k}} -\sqrt{w}\right)^2 + \left(\sqrt{h^\text{template}_{i,k}} -\sqrt{h}\right)^2
\end{equation*}
}
Then, the nearest anchor index $j'$ can be found easily from the $S_i \times S_i$ anchor grid using simple discretization, i.e., dividing the coordinates by the strides of the anchor grids.

\subsection{Training algorithm.}
The entire stochastic optimization with the aforementioned stochastic gradients is summarized as Algorithm~\ref{alg:train_attacker}.
\begin{algorithm}[ht]
  \caption{Multi-level Stochastic Optimization}
  \label{alg:train_attacker}
\begin{algorithmic}
  \STATE {\bfseries Input:} recursive attack Networks in Figure~\ref{fig3_computation_network},
  \STATE {\bfseries \hspace{1cm}} autonomous vehicle environment, proxy object detector $\vect{YOLO}$,  
  \STATE {\bfseries \hspace{1cm}} Replay buffers: $\mathcal{M}_\text{trajectory}$ and $\mathcal{M}_\text{transition}$.
  \STATE {\bfseries Output:} Fixed parameters.
 \STATE {\bfseries Initialize:} $ t \leftarrow 0$ ; $n \leftarrow 0$ ; initialize the parameters of the recursive attack networks.
 \REPEAT
 \REPEAT
    \STATE Generate and feed the attack into the environment and update state predictor
  \STATE $\quad \vect{a}_t \leftarrow \mu(\mathbf{h}_t)$
  \STATE $\quad \vect{w}_t \leftarrow \vect{Decoder}_1(\vect{Encoder}_1(\vect{x}_t), \vect{a}_t)$
  \STATE $\quad \vect{x}_{t+1}, \vect{s}_{t+1}, r_{t}, \text{done} \leftarrow \vect{Environment}(\vect{s}_t, \vect{w}_t)$
  \STATE $\quad \mathbf{h}_{t+1} \leftarrow \vect{GRU}(\mathbf{h}_t, \vect{Encoder}_0(\vect{x}_{t+1}), \vect{a}_t)$
  \STATE Add data sample to replay buffers
  \STATE $\quad \mathcal{M}_\text{transition} \leftarrow (\mathbf{h}_t, \vect{a}_t, r_{t}, \mathbf{h}_{t+1}) $
  \STATE $\quad \mathcal{M}_\text{trajectory} \leftarrow (\mathbf{x}_t, \vect{a}_t) $
  \STATE Update the parameters with samples from $\mathcal{M}_\text{transition}$ and $\mathcal{M}_\text{transition}$ using the stochastic update in~\eqref{eq:multi-time-scale}.
  \STATE $t \leftarrow t+1$; $n \leftarrow n+1$ 
 \UNTIL{$\text{done}$ is True, i.e., the episode terminates.}
  \STATE Start the next row of $\mathcal{M}_\text{trajectory}$ for a new trajectory (episode) and pop-out the oldest trajectory row in the replay buffer.
  \STATE Reset $\vect{Environment}$ and $t \leftarrow 0$
 \UNTIL{the performance meets the requirements.}
\STATE \textbf{Fix} the parameters.
\end{algorithmic}
\end{algorithm}

\subsection{Simulation environment settings.}

\subsubsection{Environment 1: Moving an UAV away from the scene}\label{sec:secenario1}
The 1\textsuperscript{st} scenario of image attack is on a UAV moving towards a car at an intersection, as shown in Figure~\ref{fig:environment1}a. In normal operation, the UAV stops at the detected target as shown in Figure~\ref{fig:environment1}b. In this scenario, the attacker's goal is to hinder the tracking controller and eventually move the UAV away from the target. Therefore, the reward function is set to promote the desired behavior as
\begin{equation*}
    r(\vect{s}_t, \vect{a}_t) = 
    \begin{cases}
        \text{distance from the target} & \text{if done}  \\
        \text{speed of the UAV} & \text{otherwise.}
    \end{cases}
\end{equation*}
We use the following termination conditions 
\begin{equation*}
    \vect{done}(\vect{s}_t, \vect{a}_t) = 
    \begin{cases}
        \text{True} & \text{if distance from the target $>$ 40 m} \\
        \text{True} & \text{if speed of the UAV $<$ 0.1 m/s} \\
        \text{True} & \text{if the UAV collides} \\
        \text{False} & \text{otherwise}.
    \end{cases}
\end{equation*}
It can be seen from the above reward function that the best possible reward is 40 when the UAV moves away from the target greater than 40 m. And on normal tracking without image attack, the UAV moves to the target and stops at the distance of 15 m from that target. After every termination, the UAV restarts from an initial position with a bounded random position displacement, i.e., uniformly random from $[-2.5, 2.5]$ for the vertical position and from $[-5, 5]$ for the lateral position. 

\subsubsection{Environment 2: Moving an UAV to a lateral direction}
The 2\textsuperscript{nd} scenario of image attack is also on a UAV moving toward a car at an intersection. There are more cars and two people at the intersection compared to the previous scenario as shown in Figure~\ref{fig:environment2}a. The UAV moves to a car that has the greatest detection-confidence out of the 5 ground vehicles as shown in Figure~\ref{fig:environment2}b. In normal operation, the UAV stops at the target. Similar to the previous scenario, the attacker's goal is to hinder the tracking controller and eventually move the UAV to a designated lateral direction (left or right). Therefore, the reward function is set to promote the desired behavior as
\begin{equation*}
    r(\vect{s}_t, \vect{a}_t) = 
    \begin{cases}
        \text{lateral coordinate} & \text{if done} \\
        \text{lateral velocity} & \text{otherwise.}
    \end{cases}
\end{equation*} 
It can be seen from the above reward function that the desired behavior attained by maximizing the reward is to move in a direction that maximizes the lateral coordinate, i.e., moving to the left. We used the same termination conditions as Environment 1.

\begin{figure}[ht]
\vskip -.05in
\begin{center}
\centerline{\includegraphics[width=0.95\columnwidth]{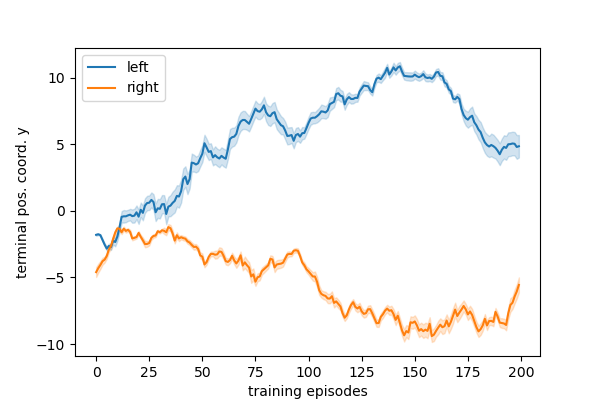}}
\caption{Effects of the two reward functions: Left vs. Right.}
\label{fig_left_vs_right}
\end{center}
\vskip -.25in
\end{figure}

\begin{figure*}[t]
\centering
\subfloat[][Scene of the street.]{\includegraphics[width=0.45\linewidth]{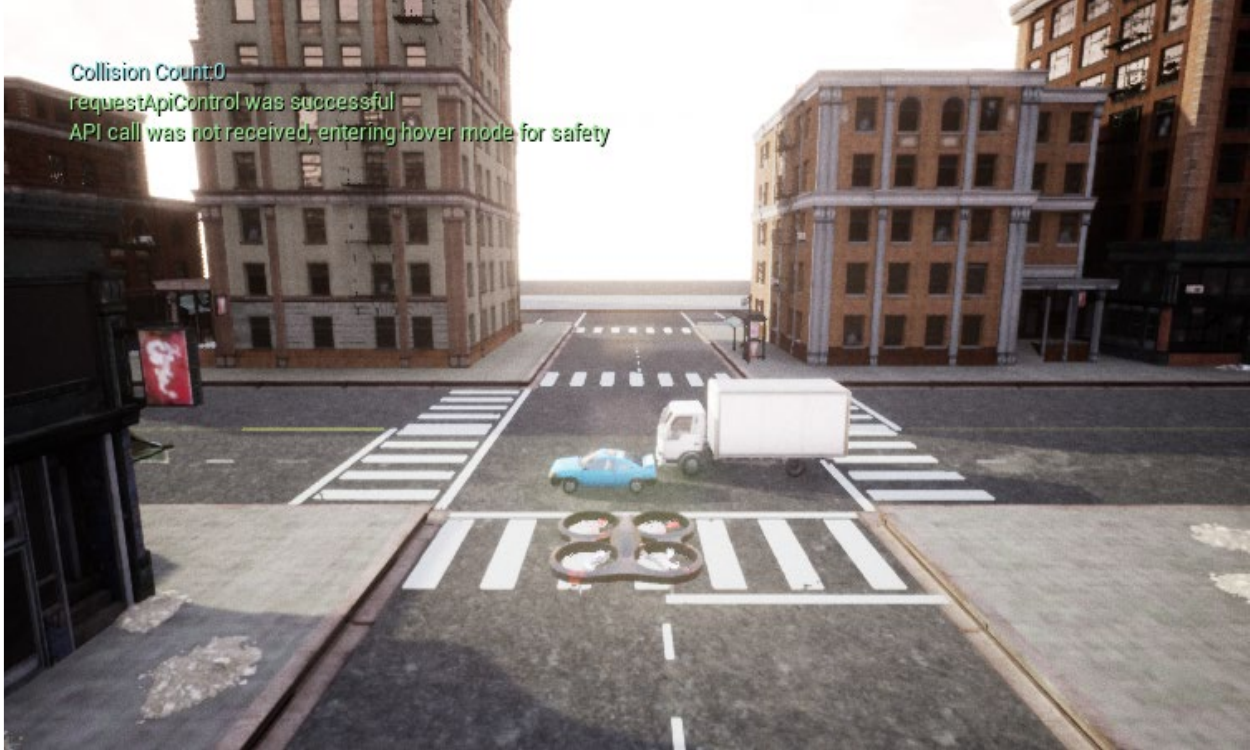}
 }
\subfloat[][Object detection.]{\includegraphics[width=0.27\linewidth]{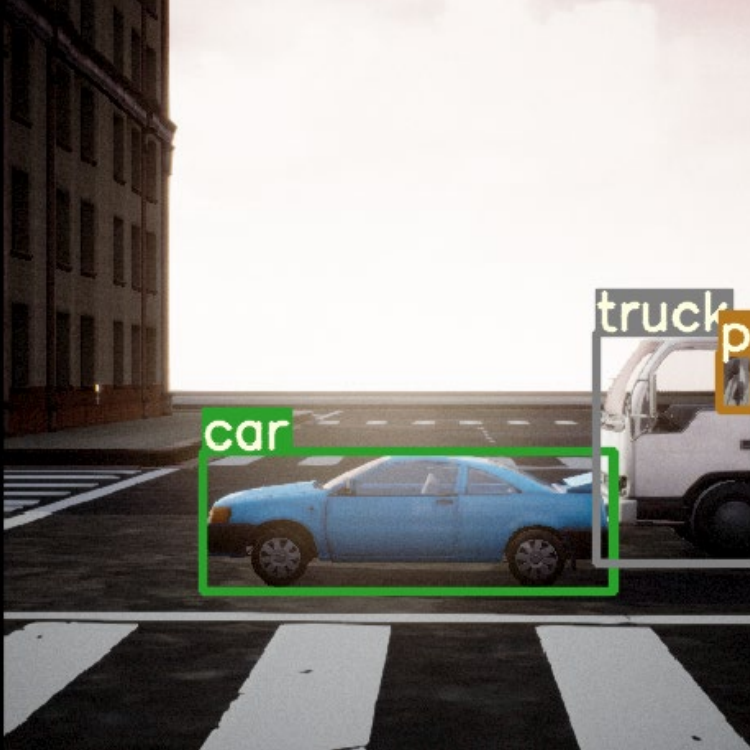}
 }
\subfloat[][Attacked detection.]{\includegraphics[width=0.27\linewidth]{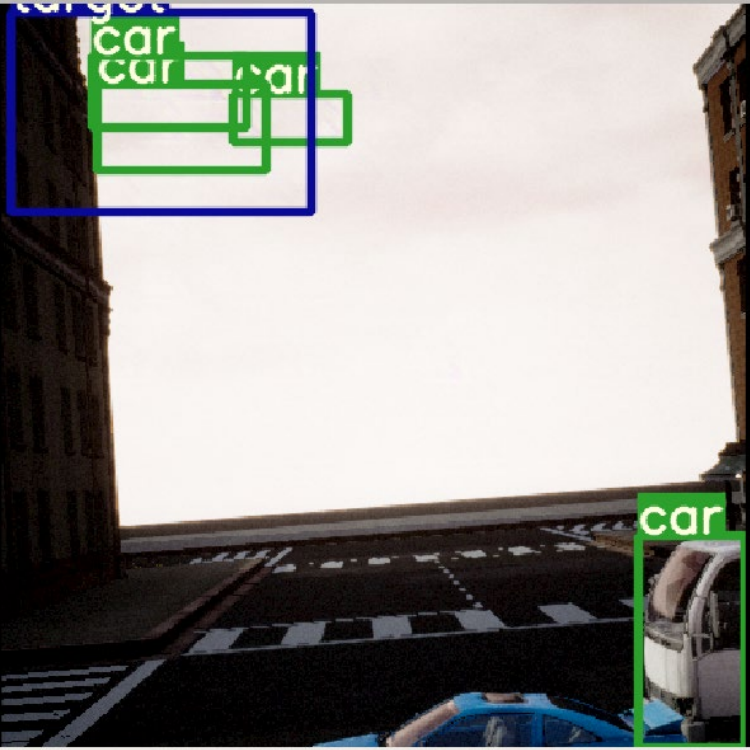}
 }
\caption{The UAV moves to the blue car as shown in (a) using the bounding box in (b). The attacker adds image perturbation to place new bounding boxes as shown in (c).} 
\label{fig:environment1}
\end{figure*}

\begin{figure*}[t]
\centering
\subfloat[][Scene of the street.]{\includegraphics[width=0.45\linewidth]{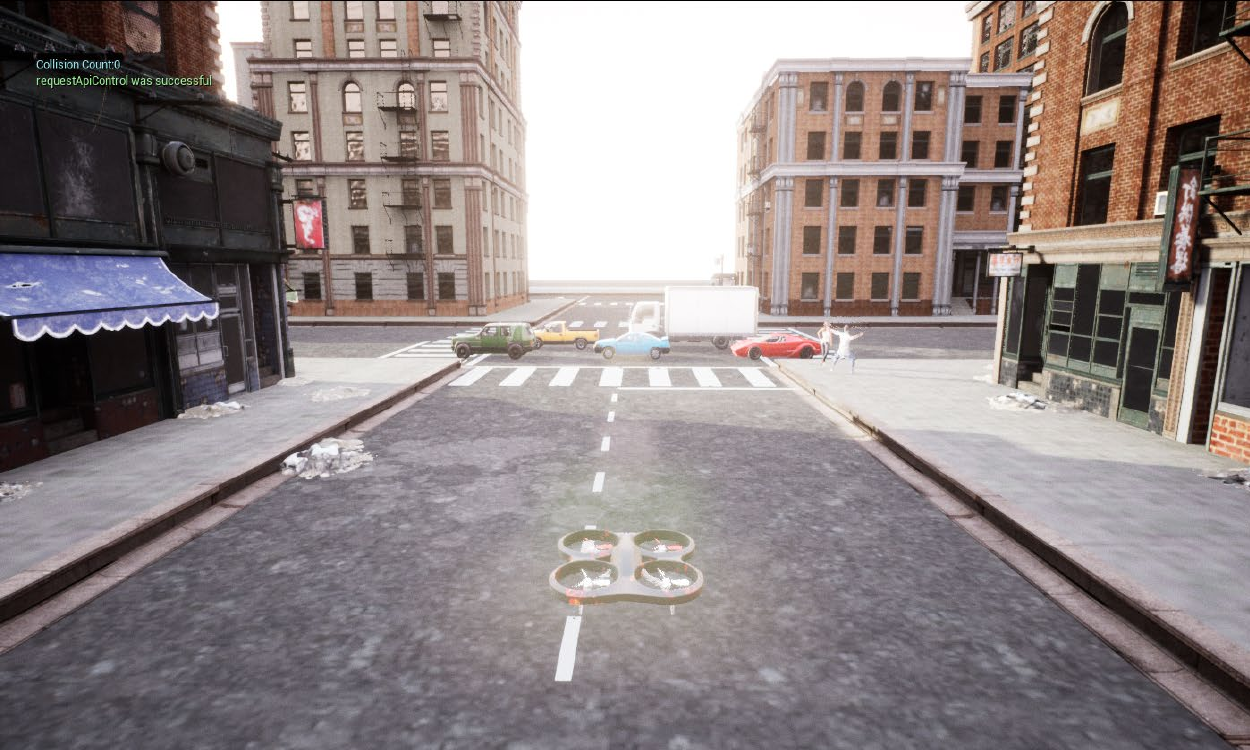}
 }
\subfloat[][Object detection.]{\includegraphics[width=0.27\linewidth]{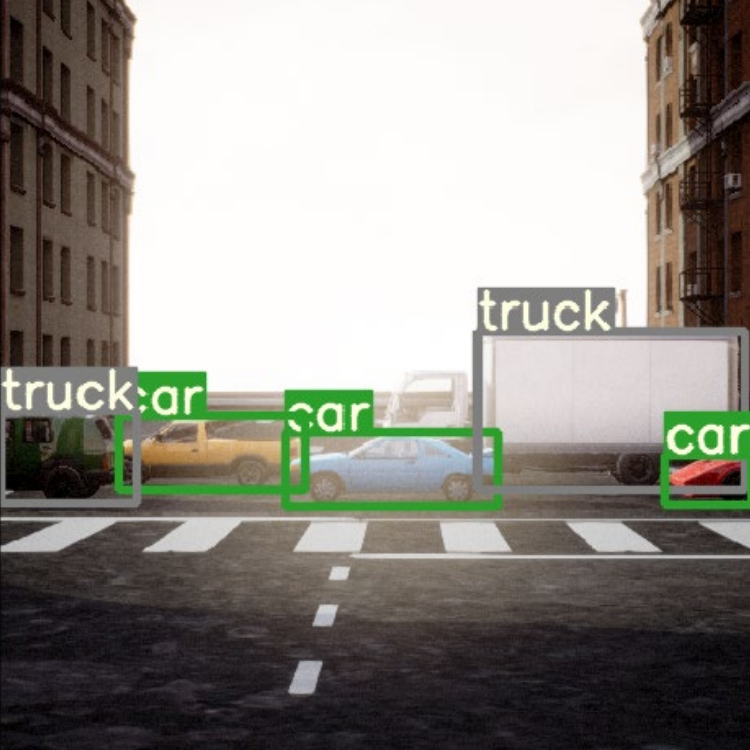}
 }
\subfloat[][Attacked detection.]{\includegraphics[width=0.27\linewidth]{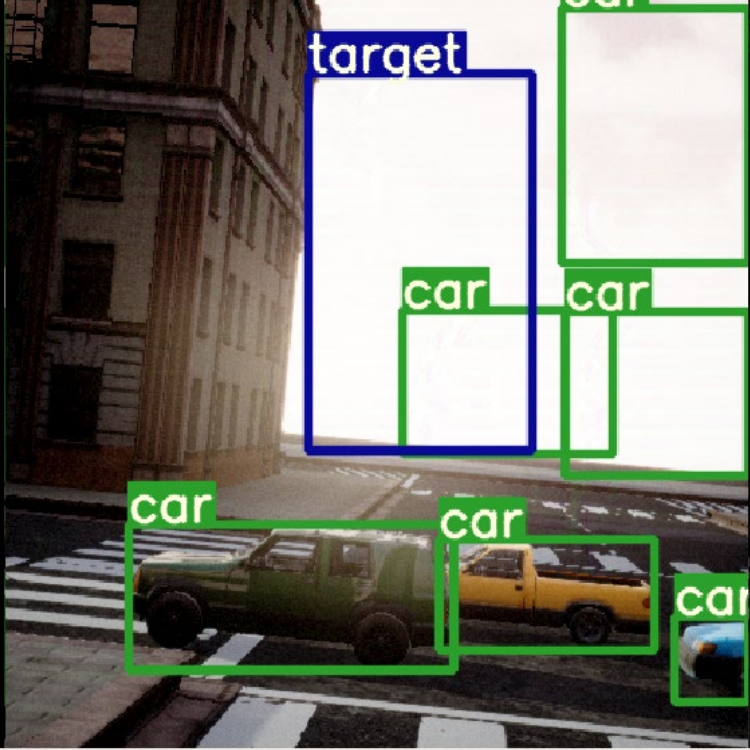}
 }
\caption{The UAV moves to a car that has the greatest detection-confidence as shown in (a) using the bounding box in (b). The attacker adds image perturbation to place new bounding boxes as shown in (c).}
\label{fig:environment2}
\end{figure*}

We tested a variation of the scenario to see whether the attacker behaves differently for different setting of reward function. Instead of making the UAV move to the left, we set the reward function to move the UAV to the right. The new reward function is as follows:
\begin{equation*}
    r(\vect{s}_t, \vect{a}_t) = 
    \begin{cases}
        -\text{lateral coordinate} & \text{if done} \\
        -\text{lateral velocity} & \text{otherwise.}
    \end{cases}
\end{equation*} 
As the learning curves show in Figure \ref{fig_left_vs_right}, the two different reward function results in different terminal positions (left vs. right). 

\subsubsection{Environment 3: Collision of a car follower using Kalman filtered object detections}
\begin{figure*}[t]
\centering
\subfloat[][Scene of the street.]{\includegraphics[width=0.45\linewidth]{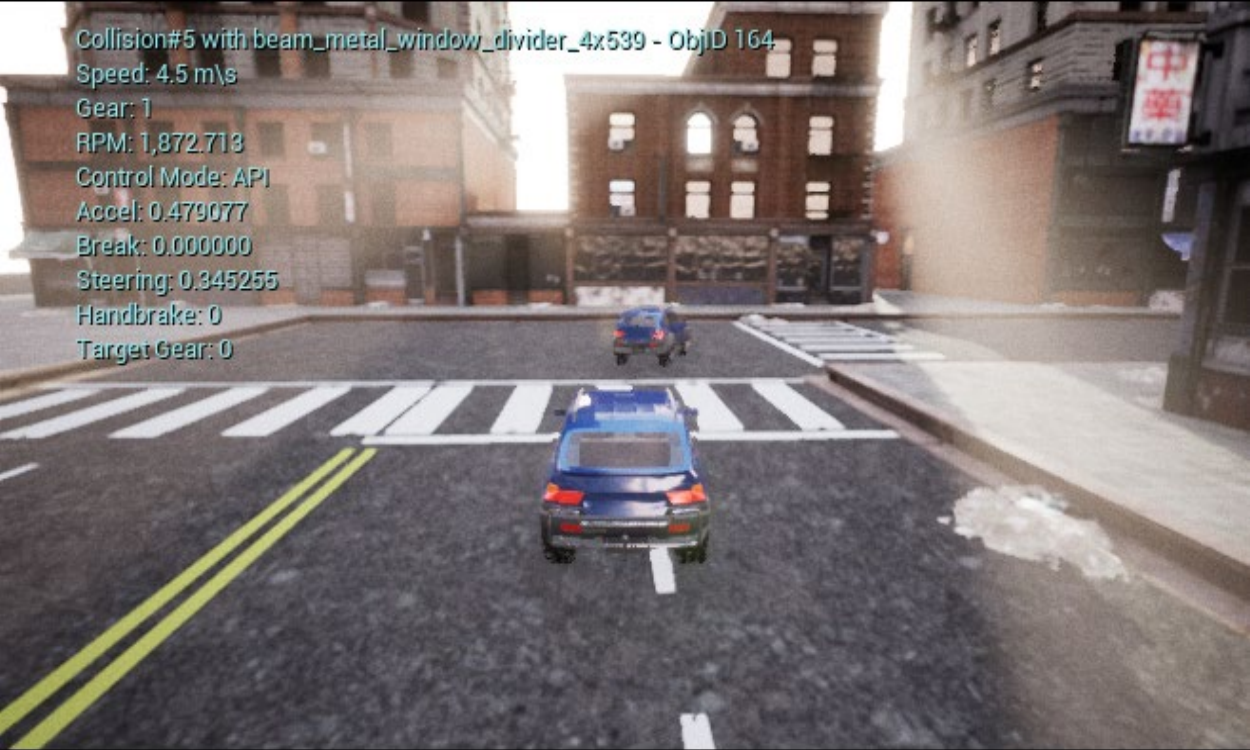}
 }
\subfloat[][Object tracking.]{\includegraphics[width=0.27\linewidth]{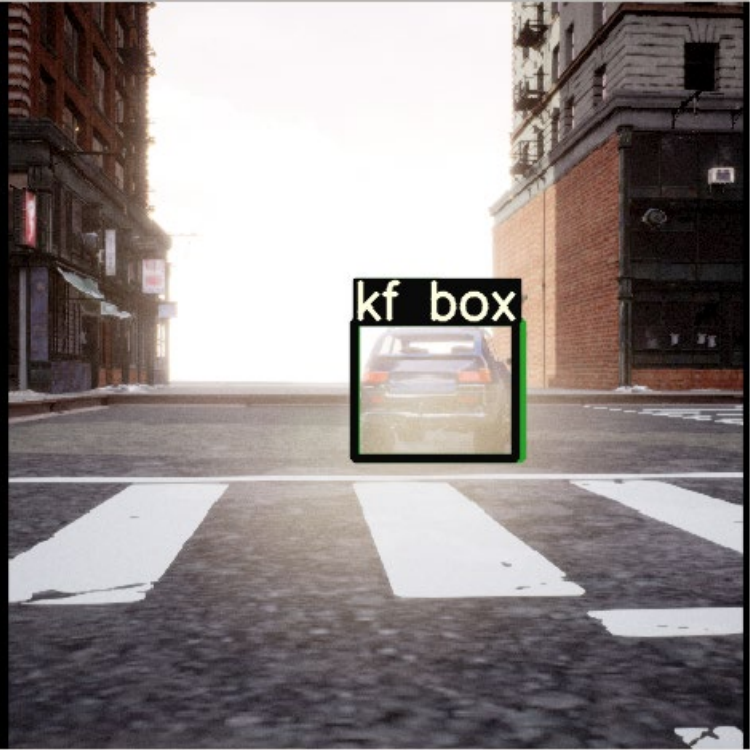}
 }
\subfloat[][Attacked detection.]{\includegraphics[width=0.27\linewidth]{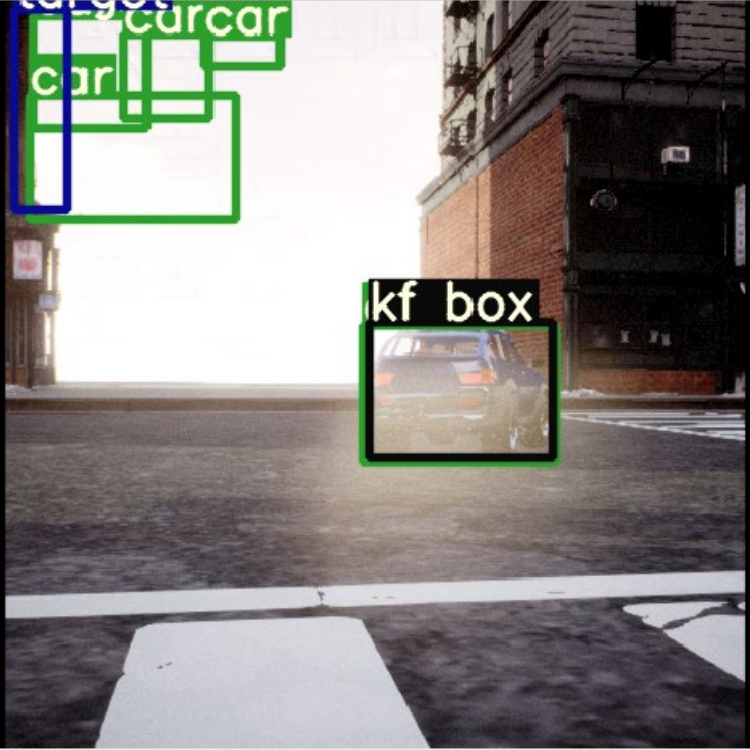}
 }
\caption{The autonomous car follows the front car as shown in (a) using a Kalman filtered bounding box denoted as \emph{kf box} in (b). The attacker adds image perturbations to place new bounding boxes as shown in (c).}
\label{fig:environment3}
\end{figure*}

In this scenario, we tested the attacker's capability with an object detection strengthened using a Kalman filter. As shown in Figure~\ref{fig:environment3}a, we simulated a car chasing scenario where the guidance system uses a filtered bounding box in Figure~\ref{fig:environment3}b using a Kalman filter (KF), similar to the KF used with an object detector in a previous work~\cite{jia2020fooling}. 
In normal operation, the autonomous car follows the car at the front.
In this scenario, the attacker's goal is to cause a collision by fooling the object tracking. For this goal, we used the following reward function as
\begin{equation*}
    r(\vect{s}_t, \vect{a}_t) = 
    \begin{cases}
        10 + \text{speed of the car} & \text{if the car collides} \\
        \text{speed of the car} & \text{otherwise.}
    \end{cases}
\end{equation*}
We also used the following termination conditions 
\begin{equation*}\label{eq:terminal_condition_car}
    \vect{done}(\vect{s}_t, \vect{a}_t) = 
    \begin{cases}
        \text{True} & \text{if the car collides} \\
        \text{True} & \text{if speed of the car $<$ 0.1 m/s} \\
        \text{False} & \text{otherwise}.
    \end{cases}
\end{equation*}

\subsubsection{Environment 4: Lost tracking of a car follower that uses multi-object tracking method}
In this scenario, we tested the attacker's capability to interrupt the tracking of the car follower. The car follower uses multi-object detection (SORT~\cite{bewley2016simple}) instead of the Kalman filter that was used in Environment 3. In normal operation, the autonomous car follows the car at the front.
In this scenario, the attacker's goal is to cause a termination of the tracking. For this goal, we used the following reward function as
\begin{equation*}
    r(\vect{s}_t, \vect{a}_t) = 
    \begin{cases}
        10 + \text{speed of the car} & \text{distance $>$ 50 m} \\
        20 + \text{speed of the car} & \text{if the car collides} \\
        \text{speed of the car} & \text{otherwise},
    \end{cases}
\end{equation*}
where the distance refers to the distance between the self-car and the front car that is being followed. 
We also used the following termination conditions 
\begin{equation*}\label{eq:terminal_condition_car}
    \vect{done}(\vect{s}_t, \vect{a}_t) = 
    \begin{cases}
        \text{True} & \text{distance $>$ 50 m}     \\
        \text{True} & \text{if the car collides}   \\
        \text{True} & \text{if time steps $>$ 700} \\
        \text{False} & \text{otherwise}.
    \end{cases}
\end{equation*}

\begin{figure*}[t]
\centering
\subfloat[][Scene of the street.]{\includegraphics[width=0.45\linewidth]{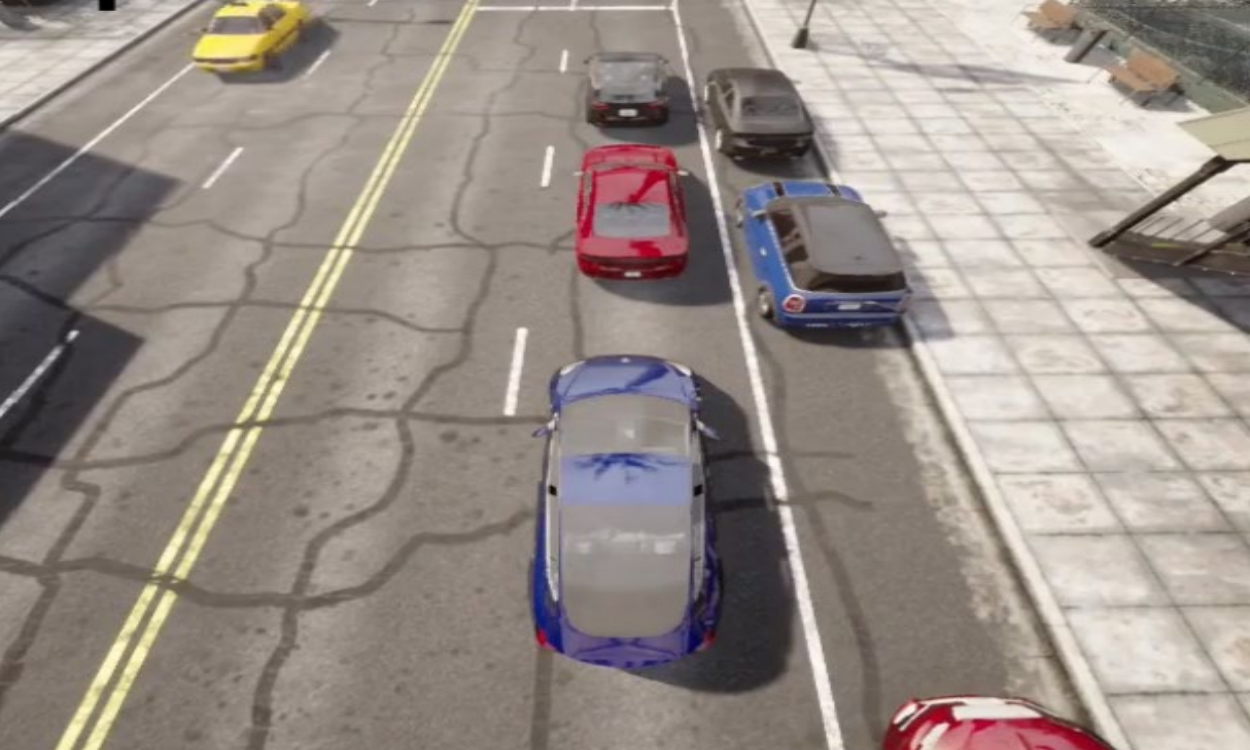}
 }
\subfloat[][Object tracking.]{\includegraphics[width=0.27\linewidth]{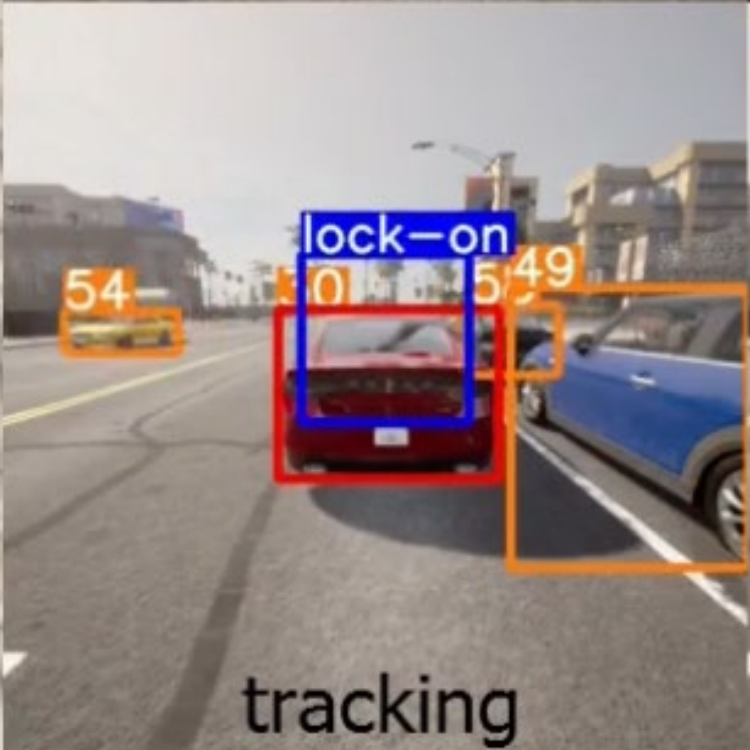}
 }
\subfloat[][Attacked detection.]{\includegraphics[width=0.27\linewidth]{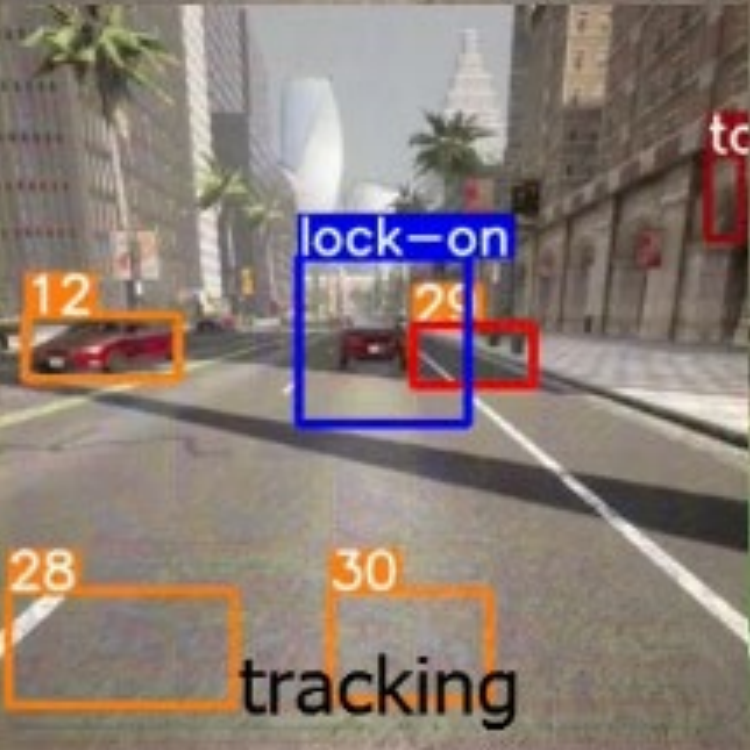}
 }
\caption{The autonomous car follows the front car as shown in (a) using the multi-object tracking (SORT~\cite{bewley2016simple}) bounding box colored red in (b). The attacker adds image perturbations to place new bounding boxes and to steer the red box as shown in (c).}
\label{fig:environment4}
\end{figure*}

\subsection{Learning Curves.}
\begin{figure*}[th]
\centering
\begin{tabular}{cc}

\subfloat[][Env. 1: \emph{Drone to a car.}]{\includegraphics[width=0.45\linewidth]{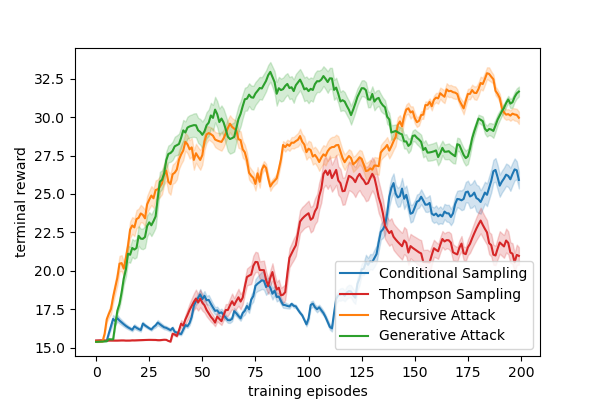}
 }
\subfloat[][Env. 2: \emph{Cars and trucks}]{\includegraphics[width=0.45\linewidth]{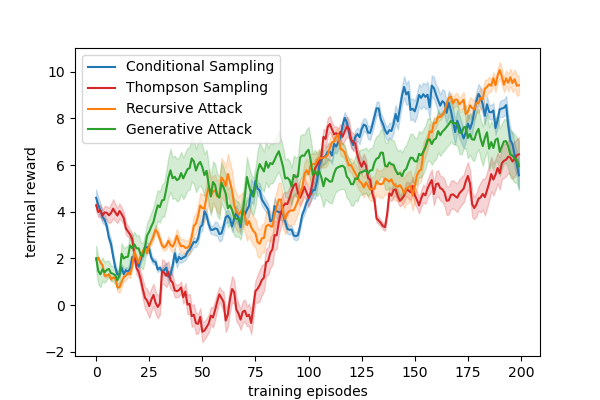}
 } \\
\subfloat[][Env. 3: \emph{Following a car.}]{\includegraphics[width=0.45\linewidth]{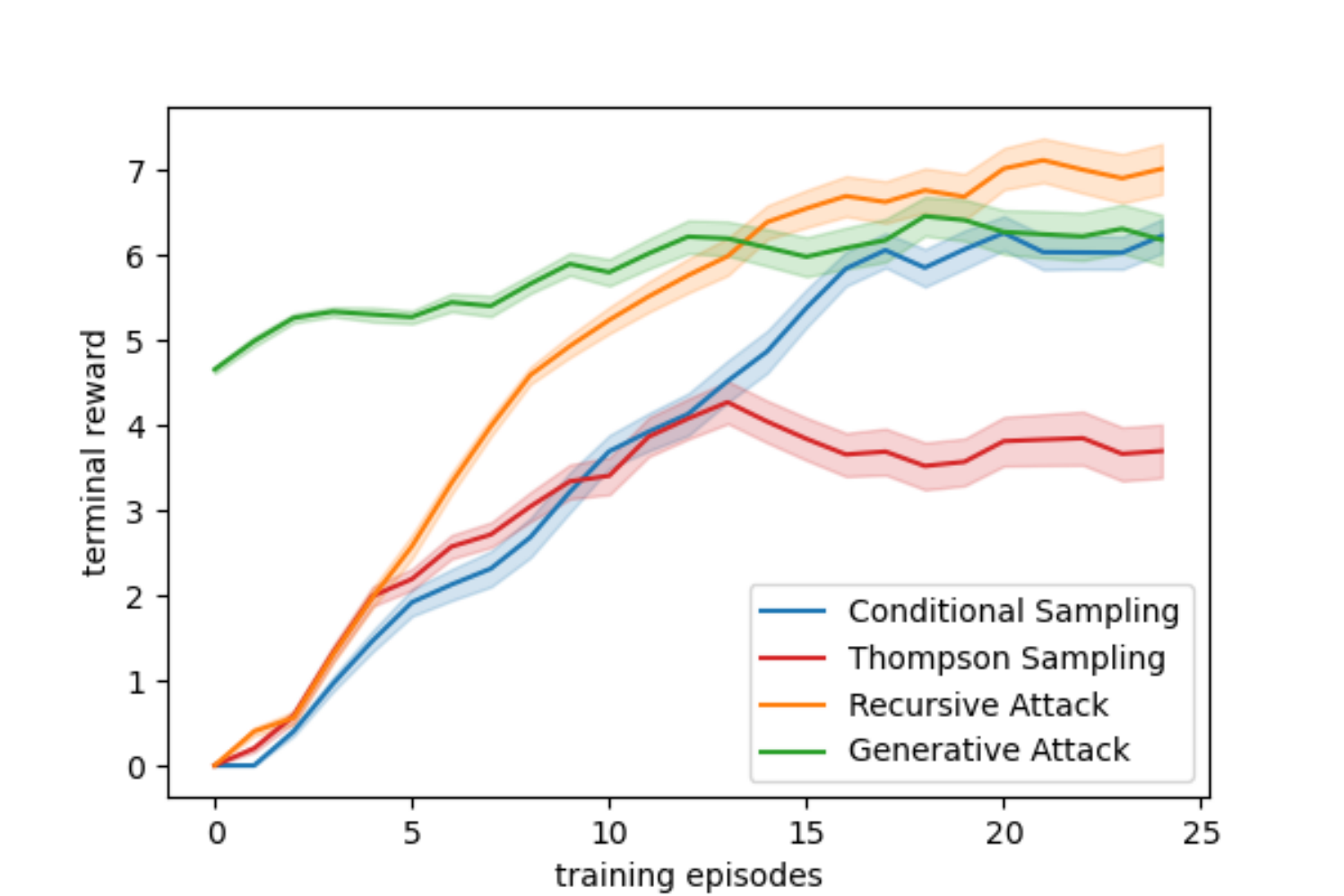}
 }
 \subfloat[][Env. 4: \emph{Following a car in traffic.}]{\includegraphics[width=0.45\linewidth]{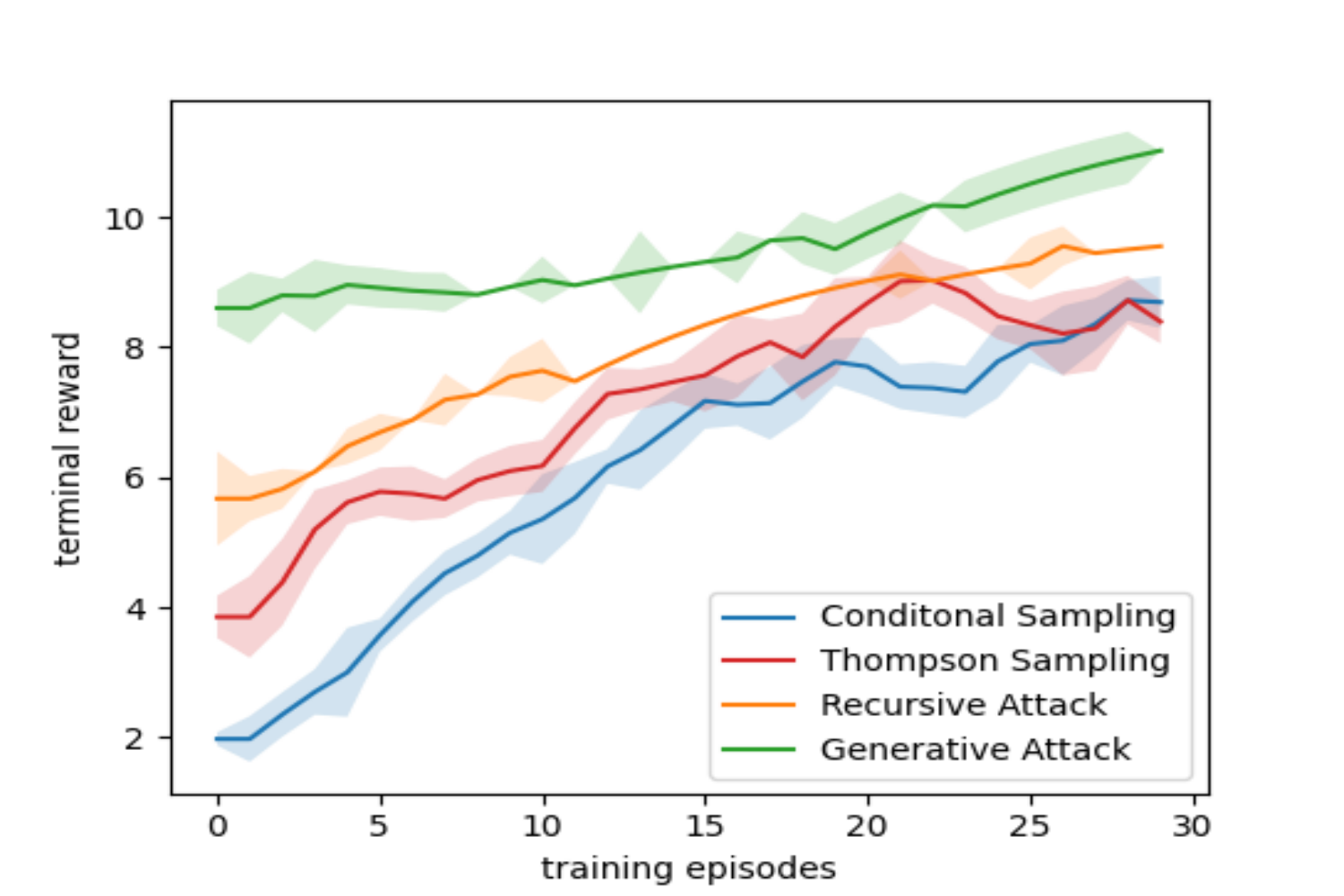}
 }
 \end{tabular}
\caption{Reinforcement Learning - Learning curves in terms of terminal rewards.}
\label{fig_learning_curve}
\end{figure*}

We ran five random reinforcement learning experiments and plotted its average and 0.1 standard deviation bounds as shaded areas shown in Figure~\ref{fig_learning_curve}a, b, and c. In the learning curves, we plotted terminal rewards since the attackers in the above scenarios aim to misguide the vehicles into certain terminal states. As we have previously seen in Table~\ref{tb:comparative_analysis}, the image attackers denoted as \emph{Recursive Attack} and \emph{Generative Attack} have greater performance. \emph{Recursive Attack} and \emph{Generative Attack} always use image attacks because they do not have attack switches, as shown in Table~\ref{tab:methods}. However, our proposed method that is to decide when to use an attack for stealthy and effective attacks. The learning curves show \emph{Conditional Sampling} has performance comparable to \emph{Recursive Attack} that always uses attacks. Especially, Figure~\ref{fig_learning_curve}b shows \emph{Conditional Sampling} has similar performance to \emph{Recursive Attack}. 

\subsection{Attack rates given image attack loss}

\begin{figure*}[th]
\centering
\begin{tabular}{cc}
     \subfloat[][Env. 1: \emph{Drone to a car}]{\includegraphics[width=0.45\linewidth]{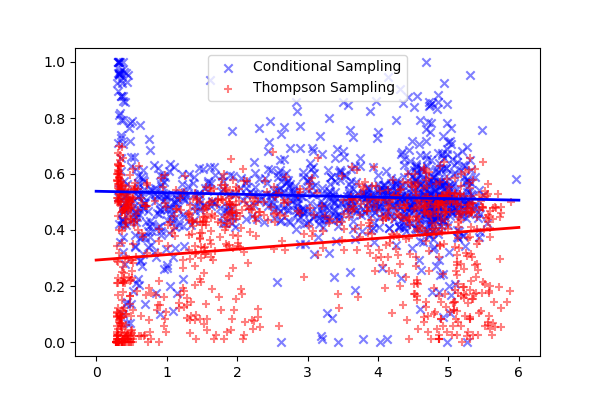}
 }
\subfloat[][Env. 2: \emph{Cars and trucks}]{\includegraphics[width=0.45\linewidth]{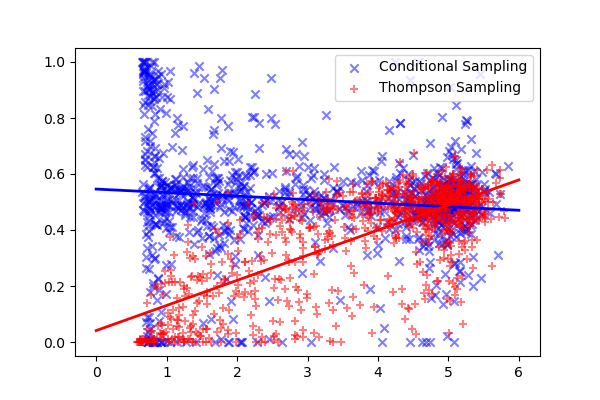}
 }\\
 
\subfloat[][Env. 3: \emph{Following a car}]{\includegraphics[width=0.45\linewidth]{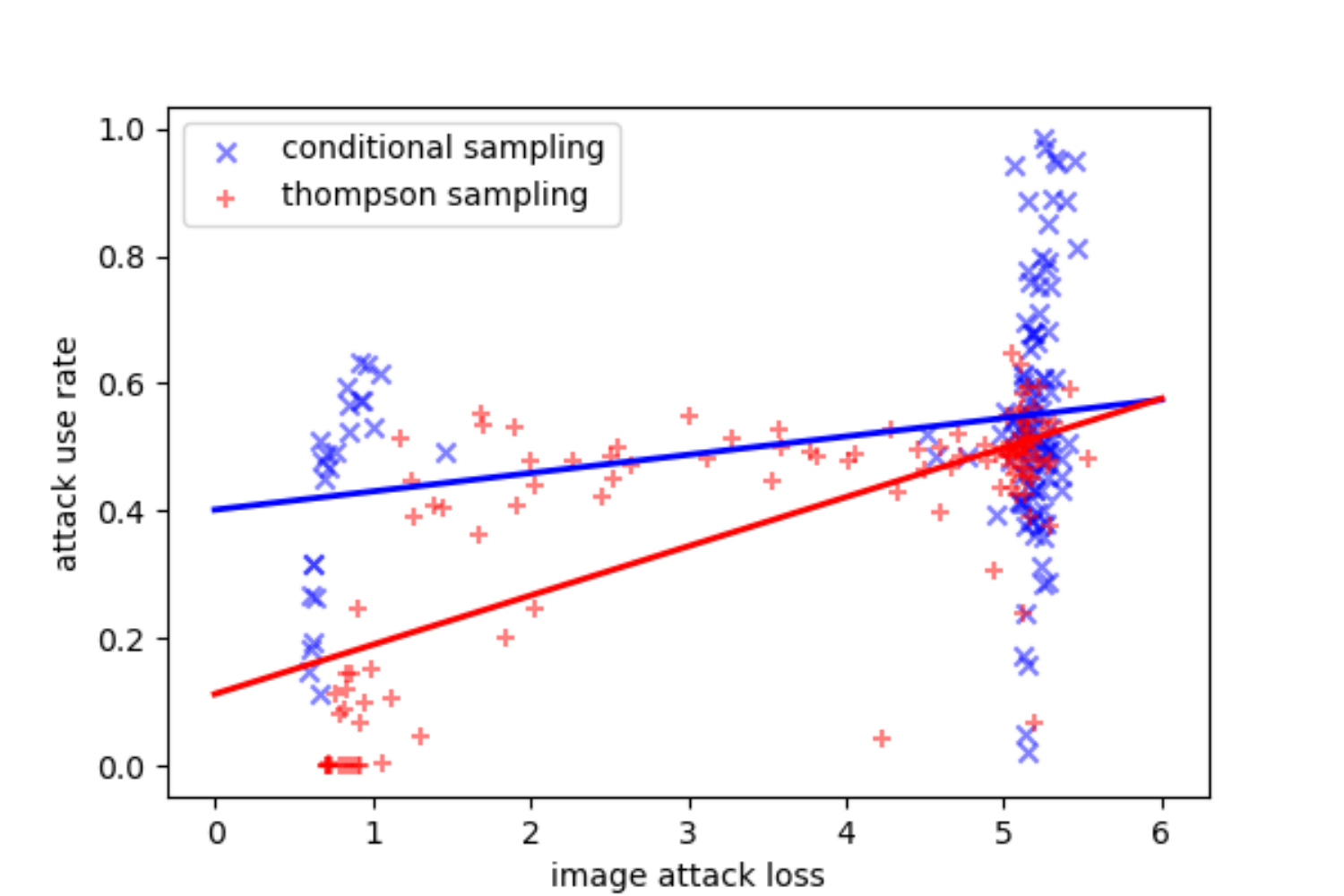}
 }
 \subfloat[][Env. 4: \emph{Following a car in traffic}]{\includegraphics[width=0.45\linewidth]{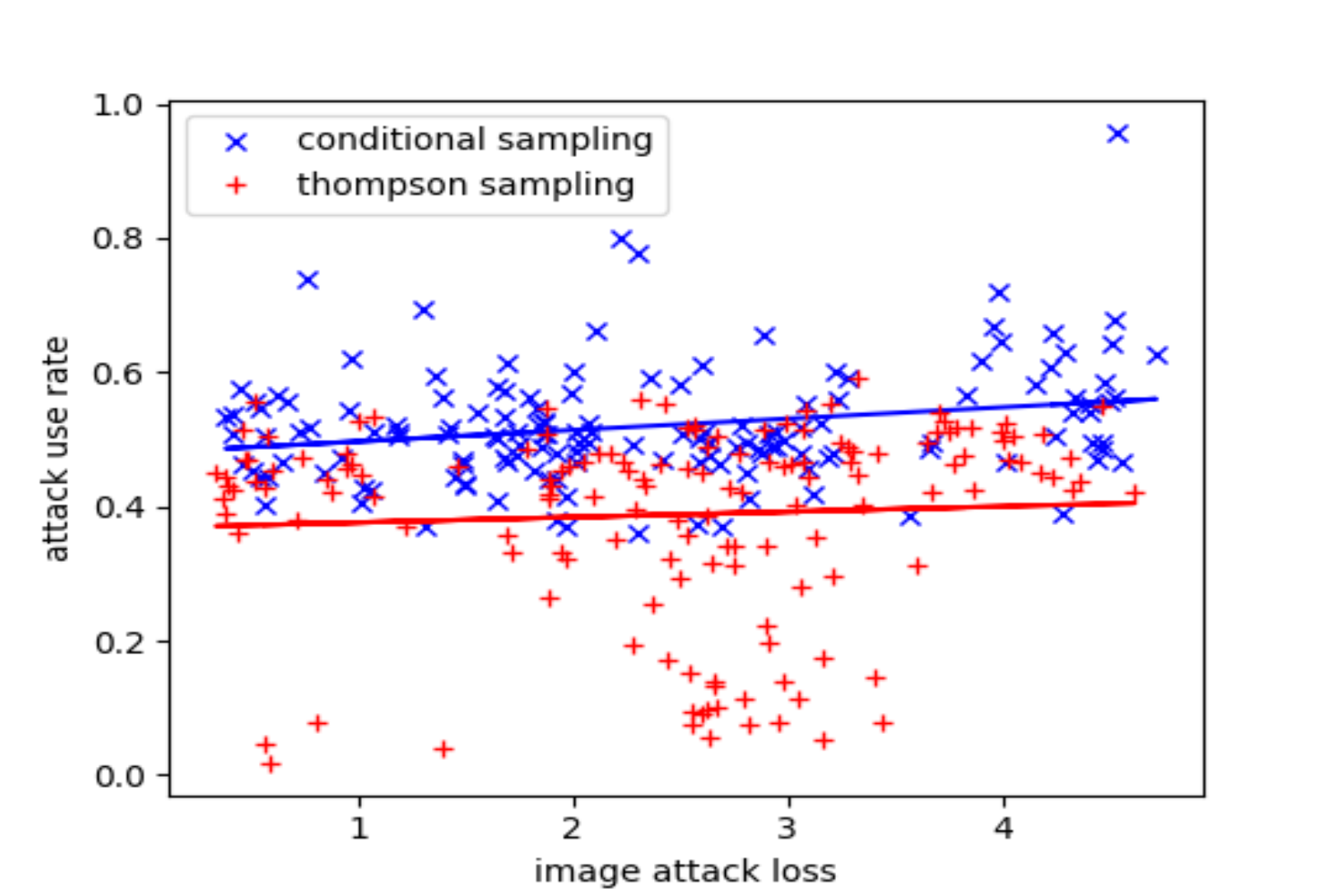}
 }
\end{tabular}
\caption{Attack rate vs. image attack loss}
\label{fig_attack_loss_corr}
\end{figure*}

The proposed methods aim to use the attack when it is good timing. We plot the attack rate vs. image attack loss using the data collected during the training. The proposed method, \emph{Conditional Sampling}, uses the attack when the attack loss is relatively low as shown in Figure~\ref{fig_attack_loss_corr}a, b, and c, compared to \emph{Thompson sampling}. In Figure~\ref{fig_attack_loss_corr}a and b, \emph{Conditional Sampling} shows negative correlations meaning that it uses more attack when the image attack loss is lower, as intended. However, \emph{Thompson Sampling} in Figure~\ref{fig_attack_loss_corr}a and b shows positive correlations that are to use more attacks when the image attack loss is higher. This unintended behavior of \emph{Thompson Sampling} could be due to the fact that the image attack loss is independent of the binary decision-making as we described in Section~\ref{sec:online_image_attack_w_switch}. The correlation coffeicents and their p-values are listed in Table~\ref{tab:correlation}. Except the Thompson sampling with the 4\textsuperscript{th} environment, the p-values are less than 0.1.

\begin{table}[h]
    \centering
    \begin{tabular}{c|c|cc}
\hline
         & \multirow{2}{*}{Methods} &correlation    &p-value      \\
         &                          &coefficient    & (2-tailed)  \\
\hline
 Env 1.                 &Conditional  Sampling    & -0.06     & 0.07        \\
\emph{Drone to a car}   &Thompson Sampling        &  0.19     & p $<$ 0.001 \\
\hline
 Env 2.                 &Conditional  Sampling    & -0.11     & p $<$ 0.001  \\
\emph{Cars and trucks}  &Thompson Sampling        &  0.73     & p $<$ 0.001 \\
\hline
 Env 3.                 &Conditional  Sampling    &  0.26     & 0.003      \\
\emph{Following a car}  &Thompson Sampling        &  0.73     & p $<$ 0.001 \\
\hline
 Env 4.                 &Conditional  Sampling    &  0.24     & 0.003      \\
\emph{Following in traffic}  &Thompson Sampling        &  0.06     & 0.45 \\
\hline
\end{tabular}
    \caption{Correlation between image attack rate and loss.}
    \label{tab:correlation}
\vskip -0.1in
\end{table}

\end{document}